%% file: ccs.tex
\documentclass[sigconf]{acmart}
\input{prefix}
\newcounter{includeSecuritySection}
\AtBeginDocument{%
  }

\copyrightyear{2026}
\acmYear{2026}
\setcopyright{cc}
\setcctype{by}
\acmConference[CCS '26]{Proceedings of the 2026 ACM SIGSAC Conference on Computer and Communications Security}{November 15--19, 2026}{The Hague, Netherlands}
\acmBooktitle{Proceedings of the 2026 ACM SIGSAC Conference on Computer and Communications Security (CCS '26), November 15--19, 2026, The Hague, Netherlands}
\acmDOI{10.1145/3830454.3846700}
\acmISBN{979-8-4007-2871-6/2026/11}

\begin{document}

\title{Beyond One-Size-Fits-All: Neural Networks for Differentially Private Tabular Data Synthesis}




 



\author{Kai Chen}
\affiliation{%
  \institution{University of Virginia}
  \city{Charlottesville}
  \country{USA}}
\email{kaichen@virginia.edu}
\author{Chen Gong}
\affiliation{%
  \institution{University of Virginia}
  \city{Charlottesville}
  \country{USA}}
\email{fzv6en@virginia.edu}
\author{Tianhao Wang}
\affiliation{%
  \institution{University of Virginia}
  \city{Charlottesville}
  \country{USA}}
\email{tianhao@virginia.edu}

\renewcommand{\shortauthors}{Kai Chen, Chen Gong, \& Tianhao Wang}


\include{abstract}







\input{core}

\input{appendix.tex}

\end{document}

%% file: prefix.tex
\usepackage{graphicx} 
\usepackage{amsthm}
\usepackage{hyperref}
\usepackage{enumitem}
\usepackage{booktabs,tabularx}
\usepackage{amsmath}
\usepackage{pifont}
\usepackage[table]{xcolor}
\usepackage{cleveref}
\usepackage{makecell}
\usepackage{multirow, multicol}
\usepackage{xspace}
\usepackage{graphicx}
\usepackage{caption}
\usepackage{subcaption}
\usepackage[linesnumbered,ruled,vlined]{algorithm2e}
\usepackage{setspace}

\newtheoremstyle{mystyle}
  {\topsep} 
  {\topsep} 
  {\normalfont} 
  {0pt} 
  {\bfseries} 
  {.} 
  {.5em} 
  {} 
\theoremstyle{mystyle}

\newtheorem{theorem}{Theorem}

\newtheorem{definition}{Definition}
\newtheorem{lemma}{Lemma}
\newtheorem{proposition}{Proposition}

\crefname{algocf}{Algorithm}{Algorithms}
\Crefname{algocf}{Algorithm}{Algorithms}

\def \privsyn{\text{PrivSyn}\xspace}
 
\def \rapp{\text{RAP++}\xspace} 

\def \aim{\text{AIM}\xspace}

\def \gem{\text{GEM}\xspace}
\def \merf{\text{DP-MERF}\xspace}
\def \ddpm{\text{DP-TabDDPM}\xspace}
\def \margnet{\text{MargNet}\xspace}
\def \ctgan{\text{DP-CTGAN}\xspace}

\newif\ifshaphered
\ifdefined\isshaphered
\shapheredtrue
\fi

\ifshaphered
\usepackage{longtable}
\newcommand\rev[1]{\textcolor{purple}{#1}}
\else
\newcommand\rev[1]{#1}
\fi

%% file: abstract.tex
\begin{abstract}
    In differentially private (DP) tabular data synthesis, the consensus is that statistical models are better than neural network (NN)-based methods. However, we argue that this conclusion does not fully capture algorithms' performance across different data regimes, particularly when correlations are densely distributed. In such complex scenarios, intricate dependencies can overwhelm statistical models, and NNs may be more competitive due to their capacity to fit complex distributions by learning directly from samples. Therefore, to fully evaluate algorithms across diverse data characteristics, we propose MargNet, which integrates adaptively selected marginals into NN training to better realize their generative capacity, and conduct evaluations on various datasets. On sparsely correlated datasets, although the prior state-of-the-art statistical baseline AIM continues to achieve high utility, MargNet achieves comparable performance with a significant speedup. On densely correlated datasets, MargNet achieves the best synthesis utility in most settings. These findings suggest that NNs remain an important approach to DP tabular data synthesis and that algorithm selection should account for dataset characteristics. We released our source code on GitHub.\footnote{\url{https://github.com/KaiChen9909/margnet}} 
\end{abstract}

%% file: core.tex
\maketitle

\section{Introduction}
\label{sec: intro}

Differential privacy (DP) for tabular data generation enables researchers and practitioners to share useful synthetic datasets with a quantifiable privacy guarantee. With the increasing research and application demands for tabular data~\cite{sangeetha2022differentially, khokhar2023differentially, barrientos2018providingaccessconfidentialresearch, Cunningham_2021}, effective methods for privately generating tabular data have received much attention~\cite{mckenna2022aim, zhang2021privsyn, vietri2022private, liu2023generating, cai2021data, tao2022benchmarkingdifferentiallyprivatesynthetic, chen2025benchmarkingdifferentiallyprivatetabular, liu2021iterative, harder2021dp}. 

Existing solutions to DP tabular data synthesis are often categorized into ``statistical methods", which usually build probabilistic models from low-dimensional statistics, and ``NN-based methods", which generate data using neural networks~\cite{chen2025benchmarkingdifferentiallyprivatetabular}.
In terms of statistical methods, AIM has been recognized as the strongest and most representative baseline on commonly used benchmarks~\cite{tao2022benchmarkingdifferentiallyprivatesynthetic, chen2025benchmarkingdifferentiallyprivatetabular}. 
\aim's success stems from two key components: adaptively selecting informative marginals under a limited privacy budget, and efficiently modeling the joint distribution with a private probabilistic graphical model (PGM)~\cite{mckenna2019graphical}. The former focuses the privacy budget on the most informative local distributions, while the latter utilizes conditional structure to approximate the full distribution from a limited set of marginals. In addition to statistical methods, various NN-based methods have been proposed to contribute to DP tabular data synthesis.
These methods broadly fall into two categories. The first applies the DP Stochastic Gradient Descent (DP-SGD) technique to train neural generators~\cite{li2025easy, xie2018differentially, tran2024differentially, chen2025benchmarkingdifferentiallyprivatetabular,castellon2023dptbarttransformerbasedautoregressivemodel,li2024privimage,gong2025dpimagebench}. The second follows a feature-DP paradigm, where privatized statistics are extracted first and then used to supervise neural training. Representative methods in this paradigm include PATE-GAN~\cite{jordon2018pate}, \merf~\cite{harder2021dp}, and \gem~\cite{liu2021iterative}, which utilize voting statistics, random Fourier feature, and marginal query to train NNs, respectively.

Various studies~\cite{mckenna2022aim,tao2022benchmarkingdifferentiallyprivatesynthetic,chen2025benchmarkingdifferentiallyprivatetabular} have evaluated prior methods and generally found AIM to be the state-of-the-art in terms of synthesis utility. 
Nevertheless, we argue that this conclusion is not complete, and a fundamental distinction between graphical models and NNs is often overlooked: \aim explicitly models the joint probability distribution of small cliques in a graph, whereas NNs typically estimate representations directly from samples. This distinction can lead to different performance across different data regimes. \rev{While \aim is frequently regarded as the state-of-the-art for utility, graphical probabilistic modeling is susceptible to numerous low-order dependencies, which can render the probabilistic graph excessively dense and intractable.} For instance, certain IoT and sensor-based datasets~\cite{pamap2} often contain attributes densely correlated with each other. Accurately tracking the data distribution in such cases necessitates modeling a larger set of marginals. Consequently, graphical models may suffer from significant information loss when simplifying the dependency structure or even computational failure. In such scenarios, NN-based generators may become more competitive because they fit complex distributions directly from samples rather than explicitly parameterizing distributions.

Recent studies have begun to investigate similar arguments. For instance, Ganev et al.~\cite{ganev2024graphicalvsdeepgenerative} reveal that certain statistical models, including MST~\cite{mckenna2021winning} and PrivBayes~\cite{zhang2017privbayes}, fail to effectively capture the data distribution in high-dimensional settings (e.g., $d \ge 128$), causing significant performance degradation. We view this evidence as suggestive, but not yet conclusive. First, the evaluated methods are not fully representative of the state-of-the-art. Specifically, the comparison overlooks more effective graphical models like \aim, as well as feature-based neural network approaches, such as \gem and \merf. Second, the analysis mainly focuses on data dimensionality and other data-independent factors, while more complex factors, such as the distribution of data correlations, have been largely overlooked. Finally, this study is primarily analytical and does not propose subsequent algorithmic enhancements, leaving room for further optimization.

Therefore, to further explore the algorithmic performance across diverse data characteristics, we propose this work. Our efforts and contributions can be summarized as follows.
\begin{enumerate}[leftmargin=*, label = \textbullet]
    \item \textbf{A New NN-based Synthesis Algorithm.} We propose MargNet, a feature-DP neural synthesis algorithm that unifies adaptive full marginal selection and differentiable marginal fitting within a single training pipeline. Unlike prior NN-based methods that rely on noisy gradient privatization~\cite{li2025easy, xie2018differentially, tran2024differentially,castellon2023dptbarttransformerbasedautoregressivemodel} or weaker feature supervision~\cite{jordon2018pate,harder2021dp,liu2021iterative}, MargNet uses privatized full marginals as adaptive and informative training targets, enabling stronger distributional supervision under the same privacy constraints. This design is intended to better realize the generative capacity of NNs for DP tabular data synthesis.

    \item \textbf{A Broader Empirical Evaluation across Data Characteristics.} We conduct a systematic comparison of statistical and NN-based synthesizers on both sparsely correlated and densely correlated datasets, covering settings with different dimensionalities and correlation structures. Beyond standard benchmark-style evaluation, our study explicitly examines how the relative performance of different paradigms changes across correlation regimes, providing a broader empirical picture than prior evaluations centered primarily on sparse-correlation datasets.

    \item \textbf{A More Nuanced Perspective on Algorithm Selection.} Our empirical results suggest that the relative strengths of statistical and NN-based synthesis methods depend strongly on dataset characteristics, rather than following a one-size-fits-all pattern. 
    \rev{More specifically, our experiments reveal that the conclusions of prior work regarding the superior performance of graphical-model-based methods are primarily derived from sparsely correlated benchmarks. Meanwhile, NN-based synthesis is especially competitive on those densely correlated regimes we study. These findings motivate algorithm selection based on data regime rather than paradigm alone.}
\end{enumerate}

\section{Preliminaries}

\subsection{DP Definition}
Differential privacy (DP) enables the extraction of aggregated statistical information while limiting the disclosure of individual information. Formally, its definition is given by:

\begin{definition}[Differential Privacy] 
\label{def:dp}
An algorithm $\mathcal{A}$ satisfies $(\varepsilon,\delta)$-differential privacy (DP) if and only if for any two neighboring datasets $D$ and $D'$ and any $T\subseteq \text{Range}(\mathcal{A})$, 
\[
\Pr[\mathcal{A}(D) \in T] \leq e^{\varepsilon}\, \Pr[\mathcal{A}(D') \in T] + \delta. 
\]
\end{definition}
\noindent Here, we say two datasets are neighboring ($D \simeq D'$) when they only differ on one tuple/sample (via adding or removing one sample). 
In practice, instead of using DP, researchers often use zero-Concentrated DP (zCDP)~\cite{bun2016concentrateddifferentialprivacysimplifications} as a tight composition tool~\cite{mckenna2022aim, zhang2021privsyn, cai2021data}, which is important in designing complex algorithms composed of multiple modules that satisfy DP. We give the definition of zCDP and its properties as follows.

\begin{definition}[zero-Concentrated DP~\cite{bun2016concentrateddifferentialprivacysimplifications}] 
\label{def:zcdp}
An algorithm $\mathcal{A}$ satisfies $\rho$-zCDP if and only if for any two neighboring datasets $D$ and $D'$ and $\alpha>1,$ $
D_{\alpha}(\mathcal{A}(D)||\mathcal{A}(D')) \leq \rho\cdot\alpha,$
where $D_{\alpha}(P||Q) = \frac{1}{\alpha-1} \ln{\mathbb{E}_{x \sim Q} \left[\frac{P(x)}{Q(x)}\right]^{\alpha}}$.
\end{definition}

zCDP has composition and post-processing properties~\cite{bun2016concentrateddifferentialprivacysimplifications}, making it a suitable choice for complex algorithm design. 
\begin{proposition}[Composition] \label{prop: compos}
    Let $f: \mathcal{D} \rightarrow \mathcal{R}_1$ satisfy $\rho_1$-zCDP and $g: \mathcal{R}_1 \times \mathcal{D} \rightarrow \mathcal{R}_2$ satisfy $\rho_2$-zCDP respectively. Then the mechanism defined as $(X,Y)$, where $X \sim f(D)$ and $Y \sim g(D, f(D))$, satisfies $(\rho_1+\rho_2)$-zCDP.
\end{proposition}

\begin{proposition}[Post-Processing] \label{prop: post-p}
     Let $f: \mathcal{D} \rightarrow \mathcal{R}_1$ satisfy $\rho$-zCDP, and $g: \mathcal{R}_1 \rightarrow \mathcal{R}_2$ is an arbitrary randomized mapping. Then $g\circ f$ is also $\rho$-zCDP.
\end{proposition}

Moreover, a $\rho$-zCDP guarantee can easily be converted to a $(\varepsilon, \delta)$-DP guarantee via \Cref{prop: zcdp2dp} \cite{bun2016concentrateddifferentialprivacysimplifications}. 
\begin{proposition} \label{prop: zcdp2dp}
    If $f$ is an $\rho$-zCDP mechanism, then it also satisfies $(\varepsilon, \delta)$-DP for $\forall \varepsilon > 0$ and, 
    \[
    \delta = \min_{\alpha>1}\frac{\exp((\alpha-1)(\alpha\rho - \varepsilon))}{\alpha-1} \left( 1 - \frac{1}{\alpha}\right)^\alpha.
    \]
\end{proposition}
\rev{In this work, we adopt zCDP for privacy accounting, primarily to maintain consistency with prior work~\cite{zhang2021privsyn,mckenna2022aim,liu2021iterative} that follows the same convention. We acknowledge that zCDP may provide less tight privacy bounds and offer less flexibility than more refined accounting frameworks in certain settings.}

\subsection{DP Mechanisms}

To achieve DP in different scenarios, various mechanisms have been proposed, such as the Gaussian mechanism~\cite{bun2016concentrateddifferentialprivacysimplifications} and the exponential mechanism~\cite{dwork2014algorithmic}. We briefly introduce them in this section. Before introducing DP mechanisms, we first need to define sensitivity: 
\begin{definition}[Sensitivity]
    Let $f: \mathcal{D} \rightarrow \mathcal{R}^k$ be a vector-valued function, then the $\ell_2$ sensitivity of $f$ is defined as,
    \[
        \Delta_f = \max_{D \simeq D'} \left\lVert f(D) - f(D') \right\rVert_2
    \]
\end{definition}

\noindent \textbf{Gaussian Mechanism}. The Gaussian Mechanism (GM)~\cite{bun2016concentrateddifferentialprivacysimplifications}, which directly adds Gaussian noise to function results, has been widely used to achieve $\rho$-zCDP. 
Specifically, let $f$ be a vector-valued function of the input data. The GM adds i.i.d. Gaussian noise with scale $\sigma \Delta_f$ to each entry of $f$,
\begin{equation}\label{eq:gaussian}
    \mathcal{A}(D) = f(D) + \sigma \Delta_f \mathcal{N}\left(0, \; \mathbb{I}\right),
\end{equation}
where $\mathcal{N}\left(0, \; \mathbb{I}\right)$ refers to the standard Gaussian distribution. The zCDP guarantee of GM is given by the \cref{prop: gm}. 
\begin{proposition} \label{prop: gm}
    The Gaussian Mechanism defined in \Cref{eq:gaussian} satisfies $\frac{1}{2 \sigma^2}$-zCDP.
\end{proposition}

\noindent \textbf{Exponential Mechanism}. Let $q_i$ be a score function for all $i \in C$. The exponential mechanism (EM)~\cite{mckenna2022aim, liu2021iterative, dwork2014algorithmic} selects a candidate $r$ according to the following distribution:
\begin{equation}\label{eq:exp mech}
    \Pr[\mathcal{A}(D) = r] \propto \exp\left(\frac{\epsilon}{2\Delta}\cdot q_r(D)\right),
\end{equation}
where $\Delta = \max_{i \in C} \Delta(q_i)$. The zCDP guarantee of EM is provided by \cref{prop: em}

\begin{proposition} \label{prop: em}
    The Exponential Mechanism defined in \Cref{eq:exp mech} satisfies $\frac{\epsilon^2}{8}$-zCDP.
\end{proposition}

\subsection{DP Tabular Data Synthesis}
\label{subsec: dp tab syn}

Firstly, we introduce some key concepts of DP tabular data synthesis necessary for later introduction. 
\begin{definition}[Tabular Dataset]
    A tabular dataset $D$ is defined as a set of $N$ record tuples $\{x_1, \ldots, x_N\}$. Each record contains $d$ values, representing its entries across $d$ attributes $\{A_1, \ldots, A_d\}$.
\end{definition}
Generally, we assume that the set $\Omega_i$ of possible values of each attribute $A_i$, namely the domain of this attribute, is public. The domain for the dataset is defined by $\Omega = \Omega_1 \times \ldots \times \Omega_d$. Another important concept is the marginal, defined as follows.

\begin{definition}[Marginal]\label{def: marginal}
    Let $r \subseteq \{1, \ldots, d\}$ be an attribute subset. We denote $D_r$ as the projection of $D$ onto the domain $\Omega_r = \Pi_{i \in r}\Omega_i$. A marginal $M$ on $r$ is a vector of frequency counts for each value in $\Omega_r$. More formally, we define it as 
    \begin{equation}\label{eq: marginal}
        M^r = \left(\sum_i \mathbb{I}\left\{x_i^r = \Omega_r[1]\right\}, \cdots, \sum_i \mathbb{I}\left\{x_i^r = \Omega_r[|M|]\right\}\right)
    \end{equation}
    where $x_i^r$ is the $i$-th record in $D_r$.
\end{definition}

Marginals are widely applied in statistical methods~\cite{cai2021data, zhang2017privbayes,zhang2021privsyn,mckenna2022aim}, and are proven effective in tabular data representation. Then, we define the DP tabular data synthesis task.

\begin{definition}[DP Tabular Data Synthesis] \label{def: dp tab syn}
    DP tabular data synthesis is defined as the problem of synthesizing an artificial tabular dataset $D'$ similar to the sensitive dataset $D$ with a DP guarantee. The two datasets are considered similar if for any $r$-way marginals, where $r\in\{1, \ldots, d\}$, the distance (e.g., $\ell_1$ distance) between the marginal from $D'$ and its counterpart from $D$ should be minimized. 
\end{definition}



{
\begin{table}[t]
\centering
\caption{The related work for DP tabular data synthesis. The model column contains the generative model used for data generation; DP mechanism refers to how the algorithm involves DP into it; the feature column includes the data feature used to capture distribution information (if used).}
\label{tab: algo}
\resizebox{\columnwidth}{!}{
    \begin{tabular}{l|ccc}
    \toprule
    \textbf{Algorithm} & \textbf{Model} & \textbf{DP Mechanism} & \textbf{Feature} \\
    \midrule 
    \aim (SOTA)~\cite{mckenna2022aim} & PGM~\cite{mckenna2019graphical} & Feature DP & Marginal \\
    \midrule
    DPGAN~\cite{xie2018differentially} & GAN & Gradient DP & - \\
    dp-GAN~\cite{zhang2018differentially} & GAN& Gradient DP & - \\
    GS-WGAN~\cite{chen2021gswgangradientsanitizedapproachlearning} & GAN& Gradient DP & - \\
    \ctgan~\cite{fang2022dp} & GAN & Gradient DP & - \\
    \ddpm~\cite{kotelnikov2023tabddpm} & Diffusion & Gradient DP & - \\
    SynLM~\cite{sablayrolles2023privatelygeneratingtabulardata} & LLM & Gradient DP & - \\
    DP-LLMTGen~\cite{tran2024differentially} & LLM & Gradient DP & - \\
    \midrule
    PATE-GAN~\cite{jordon2018pate} & GAN & Feature DP & \makecell{Teacher Voting} \\
    \merf~\cite{harder2021dp} & NN& Feature DP& \makecell{Fourier Feature} \\
    \gem~\cite{liu2021iterative}  & NN& Feature DP& \makecell{Marginal Query} \\
    \bottomrule
    \end{tabular}
}
\end{table}
}

\section{Related Work and Analysis}
\label{sec: exist work}

This section reviews related works on DP tabular data synthesis, focusing on the state-of-the-art \aim and NN-based approaches (summarized in \Cref{tab: algo}). By highlighting their respective limitations, we establish the motivation and guidance for our work.

\subsection{Related Work}
\label{subsec: exist work}

\noindent \textbf{\aim (PGM/MST)}. Before the introduction of \aim, we first present its synthesizer PGM~\cite{mckenna2019graphical}. Given selected marginals, PGM groups all attributes into several cliques $\{C_1, \dots, C_k\}$, each of which intersects with the previous clique on $S_i$ (where $i \in \{2,\ldots,k\}$). Then, by estimating the distribution of each clique, PGM can model the joint distribution by factorizing it across a chain of cliques:
\begin{equation*}
    \hat{p} = \Pr[C_1] \cdot \prod\nolimits_{i=2}^{k} \Pr[C_i \setminus S_i \; |\; S_i].
\end{equation*}

\noindent MST~\cite{mckenna2021winning} is a representative early algorithm based on PGM, which applies a one-shot marginal selection strategy and generates data through PGM. 
Building on MST, \aim~\cite{mckenna2022aim} improves upon the one-shot marginal selection by adopting an iterative marginal selection strategy, thereby enabling adaptability to complex scenarios, such as varying privacy budgets. Due to the efficacy of this strategy and the capability of PGMs to estimate the joint distribution using a limited number of marginals, relying on conditional independence assumptions across cliques, \aim has been consistently validated as the state-of-the-art algorithm in prior studies.

\vspace{1.0mm}
\noindent \textbf{Gradient-DP NN-based Method}. This category refers to the methods that apply DP-SGD~\cite{Abadi_2016} to train generative models. Their workflows are straightforward: training standard tabular generative models using DP-SGD to ensure privacy. Early efforts include DPGAN~\cite{xie2018differentially} and dp-GAN~\cite{zhang2018differentially}, which applied DP-SGD to GAN architectures originally designed for images but adaptable to tabular data. Subsequent work has focused on improving model architectures and training mechanisms. For instance, DP-CTGAN~\cite{fang2022dp} introduces conditional generators for better attribute distribution learning, while GS-WGAN~\cite{chen2021gswgangradientsanitizedapproachlearning} reduces noise by privatizing only aggregated critic gradients. This paradigm has also been extended to diffusion models and large language models, leading to methods such as \ddpm~\cite{chen2025benchmarkingdifferentiallyprivatetabular}, SynLM~\cite{sablayrolles2023privatelygeneratingtabulardata}, and DP-LLMTGen~\cite{tran2024differentially}. However, as LLMs' performance may be confounded by pre-training on extensive public datasets, which are unknown to us and can lead to privacy concerns and unfair comparisons, we omit further discussion of those LLM-based methods.

\vspace{1.0mm}
\noindent \textbf{Feature-DP NN-based Method}. Feature-DP NN-based methods first extract and privatize statistical features, then train models on these noisy features rather than the original data~\cite{liu2021iterative,harder2021dp,jordon2018pate}. This approach requires only a single noise addition step applied to feature values, avoiding the high-variance noise accumulation from multiple DP-SGD iterations. The general framework can be formulated as:
\begin{equation}\label{eq: dl model}
\begin{aligned}
    \min_{G \in \mathcal{G}} \;\; \sum\nolimits_{i=1}^{k} \mathcal{L}(f_i(D),\; \hat{f}_i({D}^{\text{syn}})), \; \; \text{s.t.} \; {D}^{\text{syn}} \leftarrow G
\end{aligned}
\end{equation}
Here, $\{f_1, \ldots, f_k\}$ are feature extractors mapping datasets to pre-defined feature spaces. The loss measures distances between features of the real dataset $D$ and the synthetic dataset ${D}^{\text{syn}}$, guiding the training of the generator $G$. Representative methods include PATE-GAN~\cite{jordon2018pate}, which uses noisy voting from teacher discriminators to train a student discriminator, which further guide the data generator; \merf~\cite{harder2021dp}, which constructs loss functions using random Fourier features mean embedding; and \gem~\cite{liu2021iterative} leverages iteratively selected marginal queries that provides a scalar answer to a specific local distributional question (e.g., one entry in \Cref{eq: marginal}), achieving superior performance among NN-based methods~\cite{chen2025benchmarkingdifferentiallyprivatetabular}.

{
    \begin{figure}
        \centering
        \includegraphics[width=0.94\columnwidth]{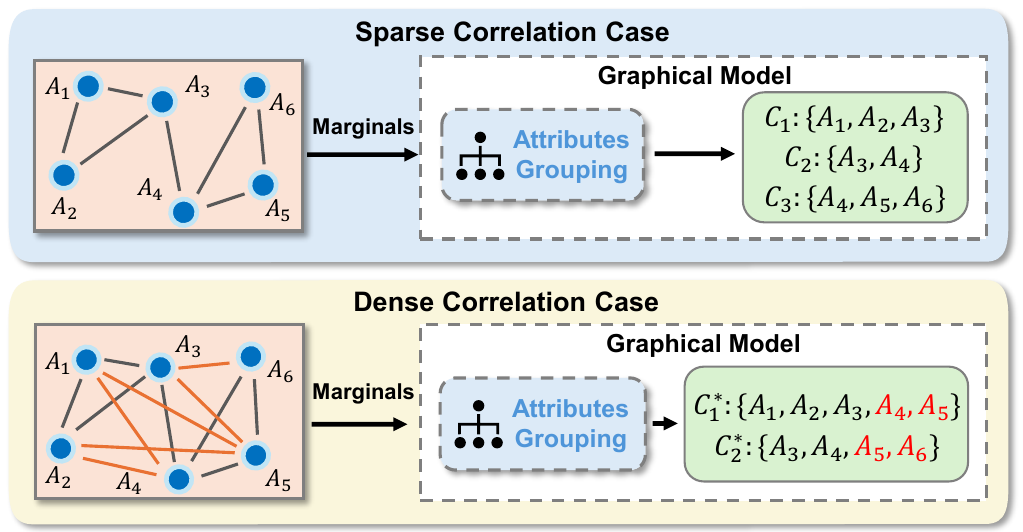}
        \caption{Examples of PGM's attributes grouping process. The blue nodes represent attributes, and the edges between nodes are marginals. The densely correlated dataset leads to cliques with larger domain sizes.}
        \label{fig: pgm case}
    \end{figure}
}

\subsection{Analysis}

\Cref{subsec: exist work} provides an overview of various solutions for DP tabular data synthesis. In this section, we conduct a deep analysis and comparison of these methods, which motivates the future algorithm design and evaluations. 

\vspace{1.0mm}
\noindent \textbf{Probabilistic Modeling v.s. Sample-based Modeling}. Although \aim's probabilistic model brings superior synthesis utility over NN-based methods~\cite{tao2022benchmarkingdifferentiallyprivatesynthetic,chen2025benchmarkingdifferentiallyprivatetabular}, prior research often overlooks the influence of correlation density on algorithmic performance. As data dependencies become denser, more correlations need to be measured and involved in models, bringing significant difficulty to modeling. \rev{It should be emphasized that dense correlation implies the existence of strong correlations among most pairwise variables. While conceptually related, this is not equivalent to higher-order correlations involving multiple variables.}



We first demonstrate the challenge that \aim faces under dense correlations by two examples provided in \Cref{fig: pgm case}. The first example considers a dataset with six attributes, where only seven marginals are identified (depicted in the upper graph of \Cref{fig: pgm case}). PGM constructs a graph model that groups all attributes into three modest cliques, each containing at most three attributes:
\[
    p \approx \Pr[C_1] \cdot \Pr[C_2 \setminus S_2 \; |\; S_2] \cdot \Pr[C_3 \setminus S_3 \; |\; S_3]
\]
Here $S_2=C_1 \cap C_2$ and $S_3=C_2 \cap C_3$. Those unmeasured marginals can be estimated through the graphical model by the conditional independence assumption, which not only reduces the complexity of fitting all correlations but also concentrates the privacy budget on those important dependencies.

However, in the second example, more strong correlations exist in this dataset, which cannot be roughly approximated by the conditional independence assumption. In this situation, PGM is forced to generate larger cliques, such as those containing four to five attributes, to accommodate these new dependencies. 
\[
    p \approx \Pr[C^*_1] \cdot \Pr[C^*_2 \setminus S^*_2 \; |\; S^*_2]
\]
Here $S^*_2=C^*_1 \cap C^*_2$.
Assuming a domain size of $100$ for each attribute, the state space of the largest clique thus increases from $10^6$ to $10^{10}$, rendering the estimation of the clique's probability distribution computationally infeasible.

Different from \aim, neural networks approximate complex data distributions by directly generating samples. The distribution of data can be derived from the output of NNs: 
\[
    p \approx F(D^{\text{syn}}),
\]
where $F(\cdot)$ is the frequency distribution function and $D^{\text{syn}}$ is the output of neural network. This estimation approach can handle densely distributed correlations by modeling $D^{\text{syn}}$, thereby enabling it to effectively address both scenarios in \Cref{fig: pgm case}.

\vspace{1.0mm}
\noindent \textbf{Drawbacks of Existing NN-based Methods}. Although our analysis suggests that NNs possess advantages over graphical models in specific situations, existing empirical evaluations~\cite{chen2025benchmarkingdifferentiallyprivatetabular,tao2022benchmarkingdifferentiallyprivatesynthetic} have rarely yielded evidence to substantiate this hypothesis. We posit that this discrepancy stems from the limitations of prior NN-based methods, which fail to fully reflect the generative potential of NNs. 

\rev{For gradient-DP methods, while they are constructed on strong generative models~\cite{xu2019modelingtabulardatausing, kotelnikov2023tabddpm}, DP-SGD introduces a trade-off between optimization quality and privacy. As the privacy budget accumulates over optimization steps, these methods often limit the training iterations, while gradient clipping and noise addition may reduce the fidelity of gradient estimates, making optimization more challenging under tight privacy budgets~\cite{chen2025benchmarkingdifferentiallyprivatetabular,tao2022benchmarkingdifferentiallyprivatesynthetic}. The feature-DP methods also have their limitations. For instance, the voting result in PATE-GAN~\cite{jordon2018pate} provides only binary supervisory signals ("yes" or "no") from the teacher ensemble, which contain less information than continuous measurements. \merf~\cite{harder2021dp} employs a random Fourier feature mean embedding, which requires infinite dimensions to capture the distribution accurately. In practice, however, only a finite number of features can be used, introducing an approximation error. Finally, although \gem~\cite{liu2021iterative} uses marginal queries to obtain stronger feature representativeness, allocates privacy budget to each selected query individually, rather than measuring the complete marginal once and reusing the resulting statistics across multiple queries. In addition, \gem follows a predetermined number of feature-selection rounds. Consequently, the algorithm does not adapt the number of selected features according to the remaining privacy budget, which may limit flexibility across datasets with different budget requirements.}

\section{\margnet}
\label{sec: margnet}

In \Cref{sec: exist work}, we identified key limitations of existing algorithm designs and their relationship with evaluation benchmarks, which motivate us to develop improved algorithms. In this section, we present \margnet, a new method designed to address this challenge.

\subsection{Algorithm Overview}

Before introducing our method in detail, we describe the design 
intuition and provide an algorithm overview. The key idea behind \margnet is to treat adaptively selected privatized full marginals as progressively refined supervision for neural generative modeling. Different from prior works~\cite{jordon2018pate,liu2021iterative,harder2021dp,xie2018differentially} that use DP-SGD to optimize the model or weak feature summaries to guide NN training, MargNet directly integrates full marginal measurements into a differentiable training pipeline. This yields a unified feature-DP methodology that couples private feature selection with neural distribution fitting, illustrated in \Cref{fig: margnet}.

{
    \begin{figure}
        \includegraphics[width=\columnwidth]{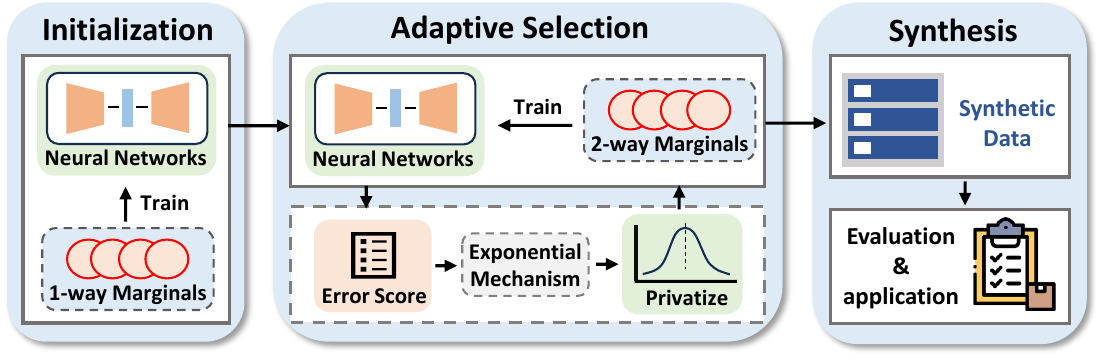}
        \caption{A brief illustration of \margnet. The model is initialized with all one-way marginals. Then, an adaptive marginal selection and model fitting step is conducted, returning a model for generation.}
        \label{fig: margnet}
    \end{figure}
}

\margnet consists of four main steps: hyperparameter initialization (line \ref{line:3-1}), model initialization (lines \ref{line:3-3}--\ref{line:3-8}), marginal selection and model update (line \ref{line:3-9}), and synthesis (line \ref{line:3-11}). First, we split the total budget into two basic units: $\rho_s$ for selection and $\rho_m$ for measurement, where $c$ is a hyperparameter constraining the maximum number of selection steps (see \Cref{appendix: set}). Then, following prior works~\cite {mckenna2022aim, zhang2021privsyn, mckenna2021winning}, we split a small proportion of privacy budget to measure all one-way marginals and fit the model as a warm-up step. After these preparations, we consider utilizing the necessary two-way marginals to capture the correlations in the dataset, which leads to further adaptive marginal selection and model update. Finally, we can generate data with the model.

This overview leaves two critical design questions unaddressed. The first is the specific strategy used for adaptively selecting the marginals. The second is the precise mechanism by which the model $G$ is trained and updated based on these selections. In \Cref{subsec: adapt frame} and \Cref{subsec: train}, we will provide a detailed description of our algorithmic design for each of these two essential steps.

\subsection{Adaptive Marginal Selection}
\label{subsec: adapt frame}

Various previous studies~\cite{mckenna2022aim, vietri2022private, cai2021data, liu2021iterative, zhang2021privsyn} have proposed iterative mechanisms to select features.
Based on these previous works, we propose an adaptive marginal selection framework more tailored to generative neural networks, as shown in \cref{algo: adapt}.

\begingroup
\setlength{\textfloatsep}{3pt}
\begin{algorithm}[!t]
\caption{MargNet}
\label{algo: margnet}
\LinesNumbered
\KwIn{Dataset $D$, total privacy budget $\rho$, hyperparameter $c$.} 
\KwOut{sampled data $D'$.} 
Set $\rho_s = 0.1\rho/c$ and $\rho_m = 0.9\rho/c$; \label{line:3-1}\\
Initialize model $G$ parameterized by $\theta$; \label{line:3-3}\\
Measure all one-way marginals $\tilde{M}^1_1, \ldots, \tilde{M}^1_d$ with budget $\rho_m$:
\[
\tilde{M}^1_i = M^1_i + \mathcal{N}\left(0, \frac{1}{\sqrt{2\rho_m}}\right);
\]\\
$S = \left\{\tilde{M}^1_1, \ldots, \tilde{M}^1_d\right\}$; \\
$\rho_{\text{total}} = \rho - d\rho_m$; \label{line:3-7}\\

Train model $G$ on $S$ by \Cref{algo: train}; \label{line:3-8}\\

Adaptively select marginals into $S$ and update model $G$ with $\rho_{\text{total}}$ by \Cref{algo: adapt};
\label{line:3-9}\\ 

Generate $D'$ from $G$; \label{line:3-11}\\
\Return $D'$.
\end{algorithm}
\endgroup

The primary workflow of \cref{algo: adapt} is an iterative process that repeatedly selects marginals and fits the model till all privacy budget is used. First, we initialize the privacy consumption accountant $\rho_{\text{used}}$ (line \ref{line:1-1}). After that, we conduct the adaptive selection. In each iteration, we first apply an exponential mechanism to select marginal $M_i$ with a commonly-used score function $q_i$~\cite{mckenna2022aim, zhang2021privsyn}:
\begin{equation} \label{eq: adapt score}
    q_i = r_i\left(\left\lVert M_i\left(G^{k-1}\right) - M_i\right\rVert_1 - n_i/\sqrt{\pi\rho_m}\right)
\end{equation}
This score can be interpreted as the expected improvement in estimating $M_i$ after we select it (estimation improvement minus noise error). The first term $\lVert M_i(G^{k-1}) - M_i(D)\rVert_1$, where $M_i(G^{k-1})$ and $M_i$ are estimated marginal by $G^{k-1}$ and real marginal, respectively, is the expected improvement we can achieve after selecting $M_i$, regardless of DP noise. However, our measurement should be privatized by DP noise. We reduce it by $n_i/\sqrt{\pi\rho_m}$, which is the expected $\ell_1$ norm of DP noise. Here $n_i$ is the size of $M_i$ and $\rho_m$ is the privacy budget for measurement; $r_i$ is a hyperparameter which is set to $1.0$ in our experiments for simplicity. The following theorem~\cite{mckenna2022aim} gives the sensitivity of $q_i$. \rev{The proof of \Cref{theorem: q} is given in \Cref{appendix: proof}.}
\begin{theorem} \label{theorem: q}
    The sensitivity of the score function $q_i$ defined in \Cref{eq: adapt score} is $r_i$.
\end{theorem}
After we obtain the newly selected marginal from the exponential mechanism, we privately measure it with Gaussian noise and train the model on all selected marginals (lines \ref{line:1-6}--\ref{line:1-8}). This part will be introduced in detail in \Cref{subsec: train}.

\begingroup
\setlength{\textfloatsep}{3pt}
\begin{algorithm}[!t]
\caption{Adaptive Selection Framework}
\label{algo: adapt}
\LinesNumbered
\KwIn{Dataset $D$, initialized model $G$, initialized marginal set $S$, privacy budget $\rho_{\text{total}}$, budgets allocated for selection and measurement $\rho_s, \rho_m$.} 
\KwOut{Generative model $G^k$.} 

$\rho_{\text{used}} = 0$, $G^0 = G$; \label{line:1-1}\\
$k=0$; \label{line:1-2} \\
\While {$\rho_{\textnormal{used}} < \rho_{\text{total}}$}{ \label{line:1-3}
    $k = k+1$; \label{line:1-4}\\
    Select two-way marginal $M_k$ using exponential mechanism with budget $\rho_s$ and score: \label{line:1-5}
    \[
    \kern 0em
    q_i = r_i\left(\left\lVert M_i\left(G^{k-1}\right) - M_i\right\rVert_1 - n_i/\sqrt{\pi\rho_m}\right)
    \]\\
    $\tilde{M}_k = M_k + \mathcal{N}\left(0, \frac{1}{\sqrt{2\rho_m}}\right)$; \label{line:1-6} \\
    $S = S \cup \{\tilde{M}_k\}$; \\
    Obtain model $G^k$ by training $G^{k-1}$ on $S$ by \Cref{algo: train}; \label{line:1-8}\\
    $\rho_{\text{used}} = \rho_{\text{used}} + \rho_s + \rho_m$; \label{line:1-9}\\
    \vspace{0.3mm}
    
    \If {$\left\lVert M_k(G^k) - M_k(G^{k-1}) \right\rVert < n_k/\sqrt{\pi \rho_m}$ \textnormal{and} $M_k$ \textnormal{is selected for the first time}\label{line:1-14}}{
        Double $\rho_m$ and $\rho_s$; \label{line:1-15}
    }
        
    \If {$\rho_{\textnormal{used}} + \rho_s + \rho_m \ge \rho_{\text{total}}$}{ \label{line:1-10}
        $\rho_s = 0.1 (\rho_{\text{total}} - \rho_{\text{used}})$; \\
        $\rho_m = 0.9 (\rho_{\text{total}} - \rho_{\text{used}});$
    }\label{line:1-12}
}
Train model $G^k$ on $S$ by \Cref{algo: train} with $\beta = $ False; \label{line:1-16}\\
\Return $G^k$. \label{line:1-17}
\end{algorithm}
\endgroup

The final step of each iteration is to update the privacy accountant and adjust the budget allocation (lines \ref{line:1-9}--\ref{line:1-12}). We first add $\rho_s$ and $\rho_m$ to $\rho_{\text{used}}$ to track cumulative consumption. Then, we update the per-iteration budget allocation $\rho_s$ and $\rho_m$ for future rounds (lines \ref{line:1-14}--\ref{line:1-15}).

This update mechanism balances two competing objectives: allocating more budget per iteration improves marginal selection reliability and reduces total iterations, while distributing budget across multiple iterations enables better noise management through Gaussian noise additivity. Prior work~\cite{mckenna2022aim} suggests that when model improvement from selecting $M_k$ falls below the expected noise error, budget increases may be beneficial:
\begin{equation}\label{eq: update}
     \left\lVert M_k\left(G^k\right) - M_k\left(G^{k-1}\right) \right\rVert_1 < n_k/\sqrt{\pi \rho_m}
\end{equation}
This condition indicates that either (1) the selected marginal itself yields limited new information, or (2) the current budget allocation produces measurements dominated by noise rather than signal. In these cases, increasing budget division can help stabilize marginal selection to choose more informative candidates and measure marginals more accurately.

We refine this into our update criterion by adding a first-time selection constraint: budget increases occur only when both \Cref{eq: update} holds and the marginal is selected for the first time. This design is based on the assumption that if a selected marginal is truly unsuitable or if measurement error is prohibitively high, these issues manifest upon first selection. By restricting budget increases to first-time selections, we avoid repeated escalation for the same marginal, instead preserving budget for additional iterations to capture more correlations, enhancing the ability to deal with complex datasets.

This budget division update step is followed by a judgment of whether the remaining budget, $\rho_{\text{total}}-\rho_{\text{used}}$, is enough for the next iteration. If not, we allocate all the remaining budget to the next round, thereby terminating the iterations because no budget remains. After the selection iterations terminate, we fit all marginals and return the model for data synthesis (lines \ref{line:1-16}--\ref{line:1-17}).

\begin{algorithm}[!t]
\caption{Marginal Training}
\label{algo: train}
\LinesNumbered
\KwIn{Model $G$ parameterized by $\theta$, iteration $T$, learning rate $r$, a set of marginals $S$, loss weight parameter $w_{1:|S|}$.}
\KwOut{Model after training $G$.} 
$t=0$; \\
$z \leftarrow \text{Random Distribution}$; \\
\While {$t < T$}{
    $D' = G(z)$; \\
    Compute loss: 
    \[
        \mathcal{L}(\theta, S) = \sum\nolimits_{i=1}^{|S|} w_i \left\lVert M_i(D') - \tilde{M}_i \right\rVert_F^2;
    \] \label{line:2-6}\\ 
    Calculate gradient: $g_{\theta} = \nabla_{\theta}\mathcal{L}(\theta, S)$; \\ 
    Update model $G$ with $g_{\theta}$ and learning rate $r$; \\
    
    $t = t + 1$;
}
\Return $G$.
\end{algorithm}

\subsection{Training with Marginals}
\label{subsec: train}

The second challenge is how to effectively train a neural network using the selected marginals. To address this, we design a differentiable training pipeline (\Cref{algo: train}), inspired by feature-fitting techniques in prior work~\cite{liu2021iterative, harder2021dp}.

The workflow of \Cref{algo: train} iteratively updates model $G$. First, we need to count the frequency of each value in the marginal distribution from the model's output while maintaining differentiability for backpropagation. To achieve this, we generate a batch of samples from $G$ where categorical values are represented as soft probability distributions, analogous to one-hot encodings but with continuous probabilities rather than discrete 0/1 values. We then average these soft probabilities across the batch to obtain the empirical frequency for each cell in the marginal. Finally, we scale these frequencies to derive the marginal estimation $M_i(D')$.

Second, we apply a weighted loss function. The weights $w_i$ are designed with two considerations: (1) since marginals are measured with varying noise levels, we assign higher weights to more reliable measurements (lower noise); (2) as we accumulate more marginals through iterations, the signal from each newly added marginal becomes diluted among an increasing number of training targets, requiring amplified weights to ensure effective learning. We formulate the weights as:
\begin{equation} \label{eq: improved weight}
w_i \propto \begin{cases}\sqrt{\rho_{m,i}},  &  M_i \text{ is not newly selected} \\d\sqrt{\rho_{m,i}},  &  M_i \text{ is newly selected}\end{cases}
\end{equation}
where $\sqrt{\rho_{m,i}}$ is inversely proportional to the noise standard deviation in measuring $M_i$. For newly selected marginals, we multiply the base weight by $d$ (the dataset dimensionality) to compensate for signal dilution as the total number of training targets grows.

\subsection{\rev{Key Improvements on Prior NN Methods}} 

\rev{In this section, we summarize the unique designs of \margnet and explain why they can help address the weaknesses of prior NN-based approaches. A brief comparison is provided in \Cref{tab: weakness}. 
\begin{enumerate}[leftmargin=*, label=\textbullet]
    \item First, we compare \margnet with \gem~\cite{liu2021iterative}, which, in most practical settings, yields better results than other NN methods in prior benchmarks~\cite{chen2025benchmarkingdifferentiallyprivatetabular,liu2021iterative}. The bottleneck of \gem stems from its fixed-iteration optimization scheme and its reliance on marginal queries to update the model. To overcome this limitation, we introduce an adaptive-iteration optimization approach in \Cref{algo: adapt} and propose a weighted full marginal optimization in \Cref{algo: train} tailored for the neural network generator, enabling the extraction of more information under the same privacy budget. 
    \item Compared with \merf~\cite{harder2021dp} and PATE-GAN~\cite{jordon2018pate}, \margnet, using full marginal features, accurately captures the frequency of every unique value within a tabular distribution, providing a more precise representation compared to statistical approximations such as random Fourier features. 
    \item Finally, \margnet overcomes the limitations of DP-SGD in the model optimization process. In DP-SGD, each optimization step consumes a portion of the privacy budget, thereby limiting the number of steps. In contrast, \margnet adopts a strategy of private measurement followed by optimization, theoretically allowing for infinite fitting iterations on the measured information.
\end{enumerate}
}

{
\begin{table}[t]
\centering
\footnotesize
\caption{The weaknesses of prior NN-based methods and the corresponding improvements of our approaches over them.}
\label{tab: weakness}
\resizebox{\columnwidth}{!}{
    \begin{tabular}{p{1.33cm}|p{3cm}|p{3cm}}
    \toprule
    \textbf{Algorithm} & \textbf{Weaknesses} & \textbf{\margnet's Design} \\
    \midrule
    GEM & $\bullet$ Marginal query is inefficient in preserving information  \par \vspace{2.5pt}
    $\bullet$ Fixed-round iteration 
    & \multirow{5}{3cm}{%
        \raggedright
        \noindent$\bullet$ Using full Marginal for more efficient information extraction \\ \vspace{2.5pt}
        \noindent$\bullet$ Adaptive-round iteration allows stronger scalability \\ \vspace{2.5pt}
        \noindent$\bullet$ Feature-level DP allows for unlimited iterations and pure gradient
    }
    \\ 
    \cmidrule{1-2}
    DP-MERF \& PATE-GAN & $\bullet$ Unrepresentative feature & \\
    \cmidrule{1-2}
    Gradient-DP \par Method & $\bullet$ limited optimization steps \par \vspace{2.5pt}
    $\bullet$ Noisy gradient & \\
    \bottomrule
    \end{tabular}
}
\end{table}
}

\subsection{Privacy Guarantee}

This subsection gives the privacy guarantee of \margnet. It is mainly based on the privacy filter~\cite{mckenna2022aim, rogers2021privacyodometersfilterspayasyougo, cesar2020boundingconcentratingtruncatingunifying, feldman2022individualprivacyaccountingrenyi}, which can guarantee DP (zCDP) under an adaptive strategy.
\begin{theorem} \label{theorem: privacy}
    Under the parameter setting used in this work, \cref{algo: margnet} satisfies $\rho$-zCDP for any $\rho>0$. 
\end{theorem}
The proof of this theorem can be found in \Cref{appendix: proof}.

\section{Theoretical Evaluation} \label{sec: theory}

Before starting our experimental evaluation, we evaluate \margnet's theoretical fitting error. We will conduct our analysis from two aspects: selected and unselected marginal errors.

\subsection{Selected Marginal Errors}

The simpler case is when a marginal is selected. Let index set $I_s$ contains indices of all selected marginals. For any $i\in I_s$, the selected marginal $M_i$ is measured as $\tilde{M}_i$, and fitted as $\hat{M}_i$ by the model. The total $\ell_2$ error can be formulated as 
\[
    \mathcal{L}_s(G) = \sum\nolimits_{i \in I_s}\lVert M_i - \hat{M}_i(G)\rVert_F^2
\]

Here, $\hat{M}_i(G)$ is achieved by first generating a batch of samples and counting the frequency of each value. This raises a problem that we are using a small number of samples to estimate a highly complex matrix, which can lead to an inherent error. Based on this intuition, we now give a lower bound on $\mathcal{L}(G)$ in the subsequent theorem. 

\begin{theorem} \label{theorem: selected lower bound}
    Assuming that the theoretically optimal estimations of selected marginals $M_1, \ldots, M_{|I_s|}$ are formulated in two-order matrices, the total fitting error of all selected two-way marginals at \margnet's any training batch has a lower bound
    \begin{equation}
        \mathcal{L}_s(G) \ge \sum\nolimits_{i \in I_s} \left( \sum\nolimits_{t=b+1}^{r(M_i)} \lambda_{i,t}^2 \right)
    \end{equation}
    Here $b$ is the batch size and $\lambda_{i,t}$ is the $t$-th largest singular value of the matrix $M_i$.
\end{theorem}

The error identified in \Cref{theorem: selected lower bound} stems from the limited sample space. This specific error is absent in PGM, as PGM directly fits the probability distributions of its cliques, ensuring their local representativeness. This highlights a fundamental trade-off: our approach simplifies the modeling to gain scalability to higher-dimensional cases, but introduces a systematic error in the process. In contrast, PGM’s more structured method avoids this sampling error, but at the cost of the computational constraints discussed previously. 

Now, we discuss an upper bound guarantee of \margnet. Deriving a deterministic upper bound is challenging, as we make no assumptions about the optimization process. Therefore, following a similar approach to the analysis of \aim~\cite{mckenna2022aim}, we instead provide a confidence bound that is conditioned on the final, observed model error with a certain probability. 

\begin{theorem} \label{theorem: selected upper bound}
    Assuming that each selected marginal $M_i$ is measured with Gaussian noise from privacy budgets $\rho_m^j$ and trained with weight $w_j$ for all $j \in J_s^i$ (here $J_s^i$ is used to demonstrate the case when a marginal is selected multiple times), we define an unbiased marginal for each $i \in I_s$ as
    \[
        \bar{M}_i \sim M_i + \mathcal{N}(0, \bar{\sigma}^2_i\mathbb{I}), \text{ where } \bar{\sigma}_i = \left(\sum\nolimits_{j\in J_s^i}\frac{w_j^2/(2\rho_m^j)}{(\sum_j w_j)^2}\right)^{1/2}
    \]
    The total fitting error of all selected two-way marginals at \margnet's any training batch has an upper bound with probability $1-\sum_i\delta_i$:
    \begin{equation} \label{eq: selected upper bound}
    \begin{aligned}
        \mathcal{L}_s(G) \leq 2 \sum_{ i \in I_s}\left(\left\lVert \bar{M}_i - \hat{M}_i(G)\right\rVert_F^2 + \sigma_i^2 F^{-1}_{\chi_{n_i}^2}(1-\delta_i) \right)
    \end{aligned}
    \end{equation}
    Here, $F^{-1}_{\chi_{n_i}^2}$ is the inverse cumulative distribution function of the Chi-squared distribution with $n_i$ degrees of freedom.
\end{theorem}

\Cref{eq: selected upper bound} consists of two terms. The first term is the discrepancy between the marginal predicted by the model and our defined unbiased estimator, which is an observable quantity at any time. The second term represents the error introduced by the DP noise. The proof of \Cref{theorem: selected upper bound} is provided in \Cref{appendix: proof}.

\subsection{Unselected Marginal Errors}

For those unselected marginals, we investigate their total $\ell_1$ estimation error as 
\[
\mathcal{L}_u(G) =\sum\nolimits_{i \in I_u} \left\lVert M_i - \hat{M}_i(G)\right\rVert_1
\]
Here we employ the $\ell_1$ error to align with the score function in the exponential mechanism (see \Cref{eq: adapt score}). This choice incurs no loss of expressiveness compared with the $\ell_2$ error discussed in the previous section, as the $\ell_1$ norm effectively captures the deviations between marginal distributions.

Previous work~\cite{mckenna2022aim} has discussed this issue under \aim's adaptive framework. The main idea of its analysis is that, even though we have not selected some marginals, we can still obtain the estimation because we expect unselected marginals to have minor errors due to the exponential mechanism. Therefore, we can obtain a bound on the unselected marginal error of \margnet from \Cref{theorem: final unselected bound}. The detailed proof is included in \Cref{appendix: proof}.

\begin{theorem} \label{theorem: final unselected bound}
In \margnet, assuming that we have conducted $K$ selection steps and selected marginal $M_\theta$ from the candidate set $C$ in the final round. The index set for unselected marginals is $I_u$. For any unselected marginal $M_i$, the estimations of it by the model $G$ and intermediate model $G^{K-1}$ are $\hat{M}_i$ and $\hat{M}^{K-1}_i$, respectively. We define an intermediate variable $B_{i, K}$ as follows.
\begin{equation} \label{eq: exp bound}
\begin{aligned}
    B_{i, K} = \frac{r_\theta}{r_i} \left\lVert M_\theta - \hat{M}^{K-1}_\theta\right\rVert_1&  + \frac{n_ir_i - n_\theta r_\theta}{r_i\sqrt{\pi \rho_s^K}} \\
    &+ \frac{\Delta_q}{r_i\sqrt{2 \rho_s^K}}\log\frac{|C|}{\delta}
\end{aligned}
\end{equation}
Here, the definition of $r_i$, $r_\theta$, $n_i$, $n_\theta$ is the same as that in \Cref{eq: adapt score}; $\rho_s^K$ is the value of $\rho_s$ in the last selection iteration; $\delta$ is a fixed parameter. Then with more than probability $1-\sum_i \delta$, we have
\begin{equation} \label{eq: final unselected bound}
    \mathcal{L}_u(G) \;\leq\; \sum\nolimits_{i \in I_u} \left( B_{i, K} + \left\lVert  \hat{M}^{K-1}_i - \hat{M}_i\right\rVert_1 \right)
\end{equation}
\end{theorem}

There are two terms in \Cref{eq: final unselected bound} that are related to the model itself. The first term is $\lVert M_\theta - \hat{M}^{K-1}_\theta\rVert_1$. We can understand it as how well we estimate $M_\theta$ before the last selection round, which is further determined by which marginals we selected before and how we fit them. The second one is $\lVert \hat{M}^{K-1}_i - \hat{M}_i\rVert_1$, which is the difference in the estimation of $M_i$ between the final model and the last model. We can assume it to be small, especially when $K$ is large, and the model converges to a stable point~\cite{mckenna2022aim}.

\section{Experiments}
\label{sec: exp}

The experimental section is organized into five parts. The first two conduct standard evaluations on sparsely and densely correlated datasets, respectively. The third part presents an in-depth comparative analysis of different algorithms' performance, followed by the fourth part quantifying the factors that can influence the algorithm performance. The final part provides variance analysis as supplementary results. 

\subsection{Sparse Correlation Experiments}
\label{sec: general comparison}

In the real world, most datasets contain sparse correlations~\cite{tao2022benchmarkingdifferentiallyprivatesynthetic,mckenna2022aim,chen2025benchmarkingdifferentiallyprivatetabular,vietri2022private,zhang2021privsyn}. Therefore, we first conduct experiments on these datasets in this section.

\subsubsection{Experimental Settings}
\label{sec: sparse setting}
We introduce our settings for sparse correlation experiments in three aspects: datasets, baselines, and metrics.

{
\begin{table}[t]
\footnotesize
    \centering
    \caption{Dataset summary. "Size" and "Dim" denote record count and dimensionality. "\#Num" and "\#Cat" represent the number of numerical and categorical attributes. The NMI refers to average normalized mutual information on all pairwise attributes, defined in \Cref{sec: exp factor}. A larger NMI indicates stronger average pairwise dependence.}
    \label{info: datasets}
    \resizebox{\columnwidth}{!}{
    \begin{tabular}{l|l|ccccc}
        \toprule
        \textbf{Type} & \textbf{Dataset} & \textbf{Size} & \textbf{Dim} & \textbf{\#Num} & \textbf{\#Cat} & \textbf{NMI}\\ 
        \midrule
        \multirow{3}{*}{\makecell[l]{Sparsely\\Correlated}} & Adult~\cite{adultdata} & 32384  & 15 & 2 & 13 & 0.05\\
        & Bank~\cite{bank_marketing_222} & 45211& 16& 6& 10 & 0.03\\
        & Loan~\cite{loandata} & 134658 & 42 & 25 & 17 & 0.03\\
        \midrule
        \multirow{4}{*}{\makecell[l]{Densely\\Correlated}} & Gauss10 & 16000  & 10 & 10 & 0 & 0.17\\
        & Gauss30 & 80000 & 30& 30& 0 &0.17\\
        & Gauss50 & 160000 & 50& 50& 0 &0.17\\
        & PAMAP2~\cite{pamap2} & 373161 & 39& 39& 0 &0.09\\
        \bottomrule
    \end{tabular}}
\end{table}
}

\noindent\textbf{Datasets}. We select three datasets, which have been widely employed by various previous works~\cite{fuentes2024jointselectionadaptivelyincorporating,chen2025benchmarkingdifferentiallyprivatetabular,vietri2020neworacleefficientalgorithmsprivate,zhang2021privsyn}, as shown in \Cref{info: datasets}. The information about the density of correlations in these datasets can be found in prior benchmark work~\cite{chen2025benchmarkingdifferentiallyprivatetabular}.

{
    \begin{figure*}[!t]
        \centering
        \includegraphics[width=\textwidth]{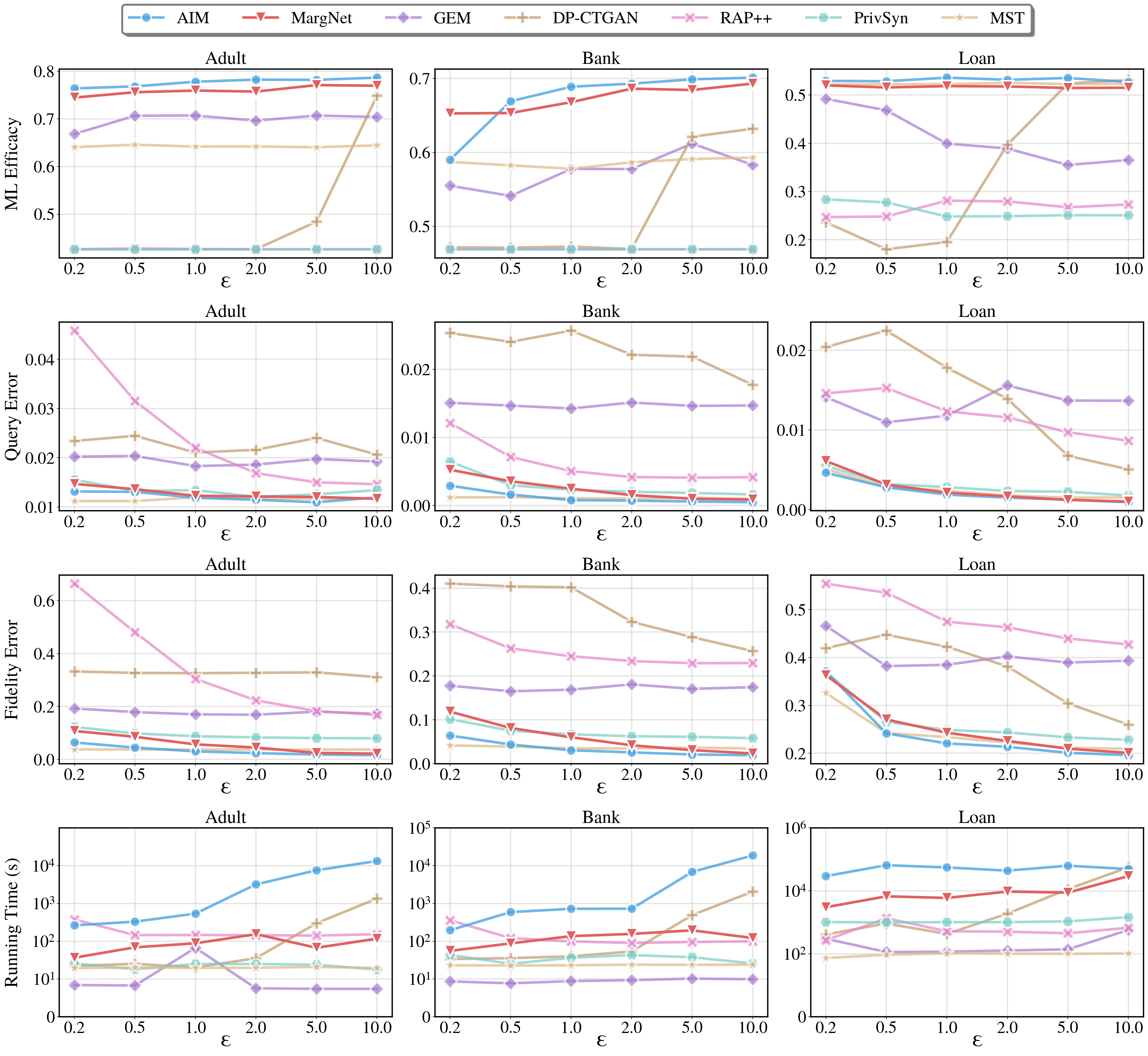}
        \caption{Evaluation results (machine learning efficacy, query error, and fidelity error) on sparsely correlated datasets. The three columns of figures depict the results on three datasets (Adult, Bank, and Loan) from left to right, respectively.}
        \label{fig: all combined}
    \end{figure*}
}

\noindent\textbf{Baselines}. For \margnet, we keep the generator structure the same as the generators of \gem and \ctgan. Detailed information about hyperparameter settings is included in \Cref{appendix: set}. We select \aim, MST, \rapp, \privsyn, \gem, \merf, \ctgan, \ddpm as our baselines. The first four methods are statistical methods. \aim is the past state-of-the-art method, and MST serves as its backend. \rapp and \privsyn are evaluated to compare with statistical methods that employ alternative synthesis mechanisms beyond PGMs. The remaining methods are NN-based methods. Some methods, such as PATE-GAN~\cite{jordon2018pate}, DP-GAN~\cite{zhang2018differentially}, have been proven ineffective~\cite{fang2022dp, harder2021dp}. Other methods like PrivMRF~\cite{cai2021data} and RAP~\cite{vietri2020neworacleefficientalgorithmsprivate} are conceptually similar to some of our baselines. Therefore, we do not include them in our experiments. Our experimental plots do not include \ddpm and \merf because of their relatively weak performance. We provide them in \Cref{appendix: sup res}.

\noindent\textbf{Metrics}. \rev{For sparsely correlated datasets, in order to keep consistency with prior works~\cite{chen2025benchmarkingdifferentiallyprivatetabular,du2024towards,zhang2021privsyn,mckenna2022aim}, we survey their experiments and focus on the most commonly used metrics. We also investigate other evaluation metrics like high-dimensional marginal (4 and 5-way marginal) query error and memory cost in \Cref{appendix: sup res}. For each algorithm, we generate five synthetic results and report the average performance over them.}
\begin{enumerate}[leftmargin=*, label=\textbullet]
    \item \emph{Machine Learning Efficacy}. We select one attribute as the label and use the other attributes as the input to train a machine learning model. The machine learning efficacy is obtained by testing this model on the real dataset. We use Catboost, XGBoost, and Random Forest models as the evaluation models, which have also been commonly used in previous works~\cite{chen2025benchmarkingdifferentiallyprivatetabular, kotelnikov2023tabddpm, du2024towards}. We report the F-1 score in our experiments because other metrics like accuracy and AUC-ROC yield similar results.
    
    \item \emph{Query Error}. The query error is obtained by comparing the results of linear queries on synthetic and real data. This metric can efficiently evaluate high-dimensional distribution similarity between synthetic and real data to test algorithms' generalization ability. Here, we conduct query error evaluation on 3-way marginals.
    
    \item \emph{Fidelity Error}. For low-dimensional distributions, directly comparing them is the most precise way for evaluation. We apply Total Variance Distance (TVD) as the fidelity error of 2-way marginals, which is also commonly used in previous benchmarks~\cite{mckenna2022aim, tao2022benchmarkingdifferentiallyprivatesynthetic, chen2025benchmarkingdifferentiallyprivatetabular}.

    \item \emph{Running Time}. Here, we utilize the execution time as the measurement of efficiency, which can determine algorithms' practicality in real-world applications but is often overlooked.
\end{enumerate}

\subsubsection{Experimental Results}

We plot the synthesis utility and time efficiency of different algorithms on three sparsely correlated datasets in \Cref{fig: all combined}.

{
    \begin{figure*}[!t]
        \centering
        \includegraphics[width=\textwidth]{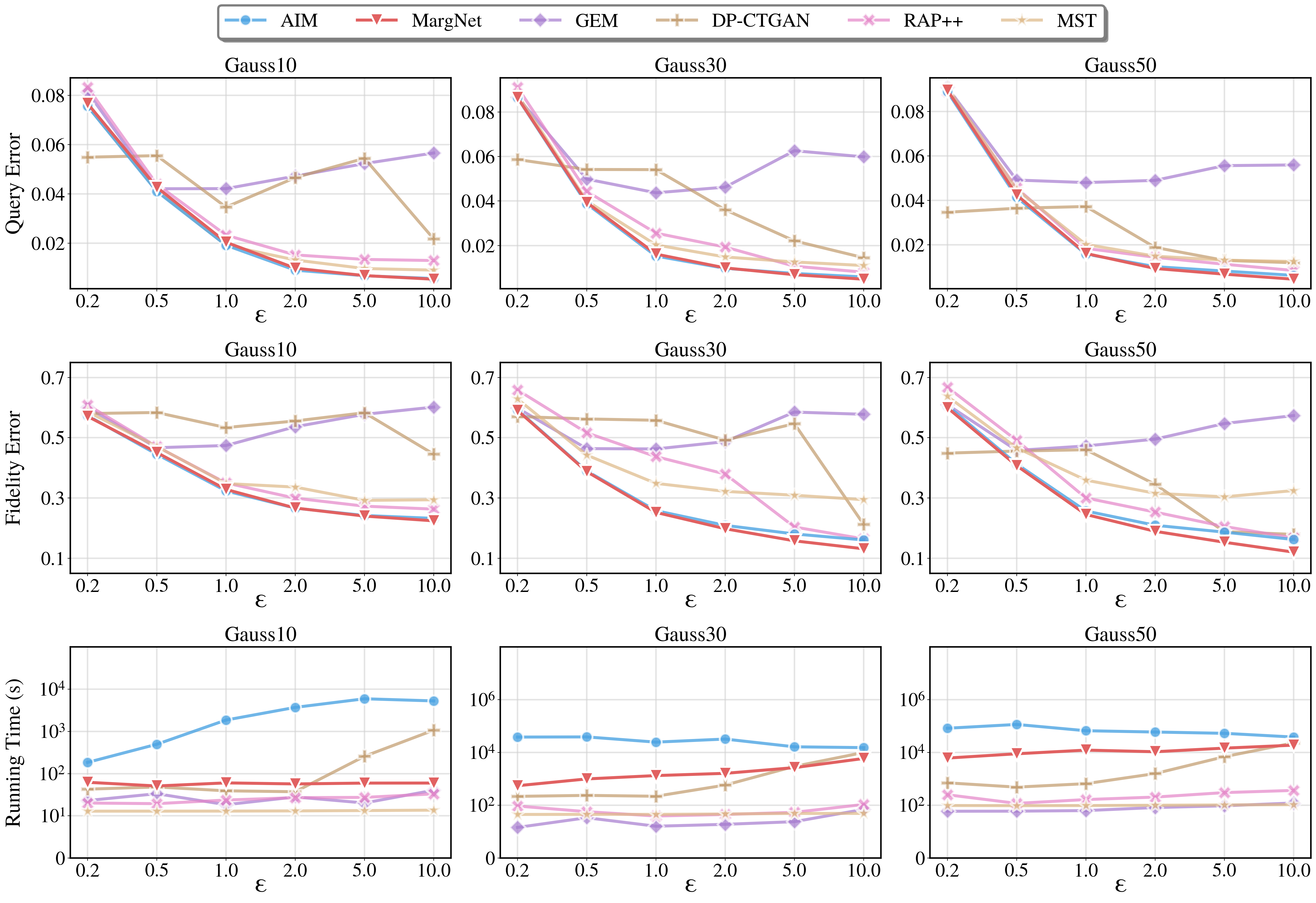}
        \caption{Different algorithms' query errors, fidelity errors, and running time on artificial densely correlated datasets. The three columns of figures depict the results on three datasets, Gauss10, Gauss30, and Gauss50 from left to right, respectively.}
        \label{fig: gauss res}
    \end{figure*}
}

\vspace{1.0mm}
\noindent \textbf{\aim still outperforms other methods in synthesis utility, but is weak in efficiency.}  Regarding utility, \aim shows superior performance on these sparsely correlated datasets, which aligns with observations in prior benchmark works~\cite{chen2025benchmarkingdifferentiallyprivatetabular,tao2022benchmarkingdifferentiallyprivatesynthetic}. The utility gap between \margnet and \aim exists but narrows as the privacy budget increases, with their performance becoming close at larger budgets (e.g., $\varepsilon = 10.0$). We attribute this to the fact that AIM's primary advantage, its efficient privacy budget utilization from conditional independence estimations, becomes less significant when the budget is enough to measure more marginals. \rev{Compared with \aim, MST, another algorithm built on PGM, is more efficient. However, this primarily stems from its non-adaptive marginal selection mechanism, resulting in degraded utility.} Furthermore, we observe that \margnet is significantly more time-efficient than \aim on these datasets, achieving an average speedup of 7$\times$.

\vspace{1.0mm}
\noindent \textbf{\margnet consistently outperforms other NN-based methods and several statistical methods in synthesis utility}. Aligning with what we claimed before, existing NN-based methods, including \ctgan and \gem (as well as \ddpm and \merf, whose results are provided in \Cref{appendix: sup res}), face challenges. Under low privacy budgets (e.g., $\varepsilon \leq 5.0$), we observe that \ctgan, which relies on DP-SGD, becomes almost ineffective. Although its utility improves with a larger budget, even at a high $\varepsilon = 10.0$ in our settings, its synthesis utility is still weaker than that of \margnet, especially in query error and fidelity error. This suggests that a general-purpose DP mechanism like DP-SGD is less effective for this task than \margnet's more specialized architecture. 

Another noteworthy experimental observation is that the performance of \gem remains largely stagnant as the privacy budget increases, consistently under-performing relative to \margnet. This lack of scalability stems from the fixed-round optimization in \gem, which restricts the model's ability to fully utilize a larger privacy budget for additional measurements. In contrast, \margnet addresses this limitation, effectively leveraging increased privacy budgets to capture more comprehensive distributional information to build the generative model.

Furthermore, \margnet demonstrates superior performance over other traditional statistical methods except \aim. Regarding MST, we observe a significant deficiency in its ML efficacy. One possible reason is its one-shot marginal selection strategy: by constructing the correlation graph in a single pass without subsequent iterative refinement, MST fails to effectively capture global correlations. As for other baselines (\rapp and \privsyn), we hypothesize that their performance gap stems from both their marginal selection mechanisms and underlying synthesizers, which are inferior to the adaptive marginal selection strategy and PGM used in \aim~\cite{chen2025benchmarkingdifferentiallyprivatetabular}. Collectively, these results support our conclusion that neural networks remain a competitive paradigm for DP tabular data generation.

{
    \begin{figure*}[!t]
        \centering
        \includegraphics[width=\textwidth]{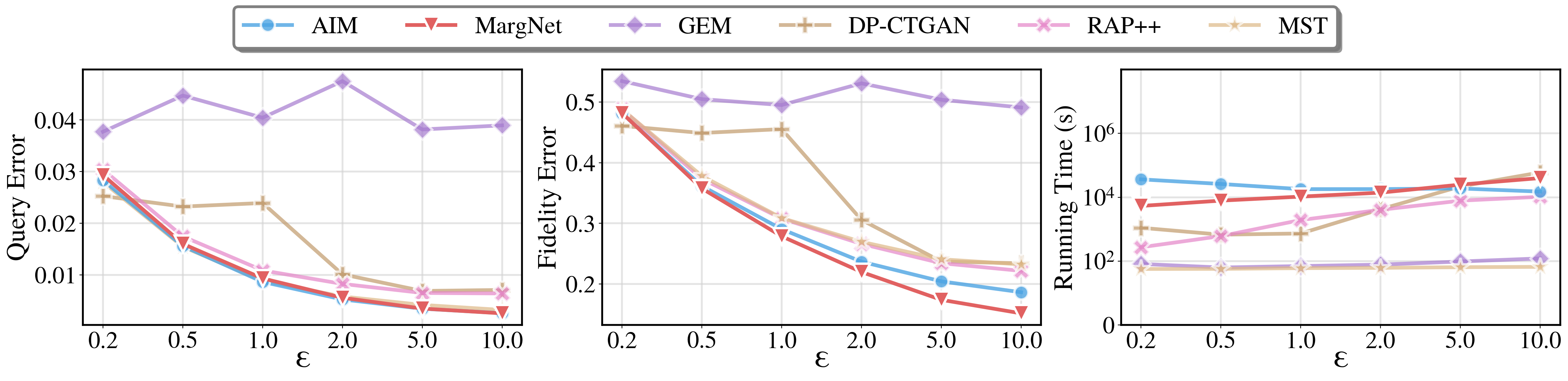}
        \caption{Different algorithms' query errors, fidelity errors, and running time on real densely correlated datasets, PAMAP2.}
        \label{fig: pamap res}
    \end{figure*}
}

\subsection{Dense Correlation Experiments} 
\label{sec: dense comparison}

Another type of dataset we consider is densely correlated data, which can be more challenging for \aim due to the larger number of dependencies. Such settings may therefore provide a regime in which NN-based methods become more competitive. Here, we focus on evaluating algorithmic performance on these datasets.

{
    \begin{figure*}[!t]
        \centering
        \includegraphics[width=\textwidth]{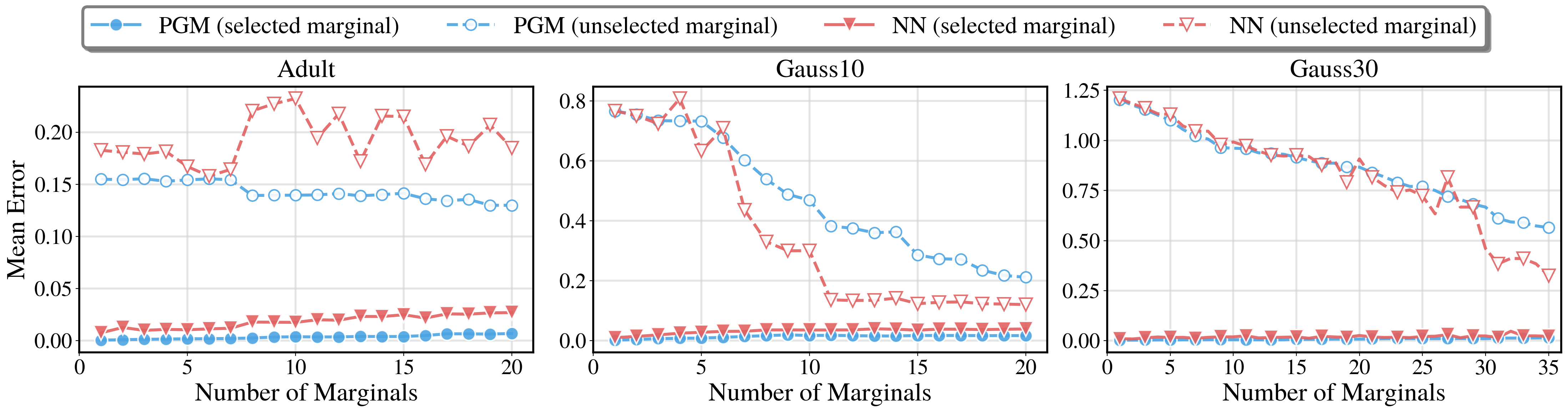}
        \caption{$\ell_1$ distance error of selected and unselected marginals fitting by PGM and NN. The marginals are iteratively and randomly selected for fair comparison.}
        \label{fig: random marginal loss}
    \end{figure*}
}

\subsubsection{Experimental Settings}
The baselines used in this section are kept the same as those in \Cref{sec: general comparison}. Therefore, we omit the description of them. Here, we introduce the datasets and metrics employed in this experimental study.

\noindent\textbf{Dataset}. We consider both artificial and real datasets to better demonstrate the algorithms' performance in densely correlated cases. For the artificial dataset, we fix a covariance matrix and generate three multi-dimensional Gaussian-distributed datasets, as shown in \Cref{info: datasets}. The pairwise correlations are set to $0.8$ so that each dataset is a densely correlated dataset. Here, the pairwise correlations are configurable. Therefore, in \Cref{sec: exp factor}, we can manually adjust them to investigate the influence of the correlation density on the algorithm's performance.

For the real dataset, we consider the PAMAP2 dataset~\cite{pamap2}. This dataset contains multi-sensor physical activity data, and due to the interconnected nature of human movement, we expect most variables to be correlated. \rev{Here, we treat the data in each timestep in PAMAP2 as one record and provide record-level DP. We filter out some columns, such as timestep, and include only 39 columns.}


\noindent\textbf{Baselines}. We consider the same algorithms as those in \Cref{sec: general comparison}. One exception is \privsyn, which has poor performance on densely correlated datasets and encounters an out-of-memory error on the PAMAP2 dataset. Therefore, we omit it in this situation. 

\noindent\textbf{Metrics}. We follow the metrics used in the assessment on sparsely correlated datasets. For Gaussian distribution data, machine learning efficacy is not necessary because a combination of all pairwise relations determines the covariance matrix. For the PAMAP2 dataset, we selectively extract the sensor data columns, as several auxiliary columns are invalid according to the documentation from the data source. Furthermore, to maintain data simplicity, we restrict our data to a single activity (all data records have the same label). Therefore, downstream machine learning performance evaluation is not required. Therefore, we consider query error and fidelity error for utility evaluation.

\subsubsection{Experimental Results}

The quantitative evaluation results on densely correlated datasets are plotted in \Cref{fig: gauss res}. 

\vspace{1.0mm}
\noindent \textbf{On densely correlated datasets, \margnet performs better than other methods, especially in high dimensions}. In these situations, \margnet and \aim are the top two algorithms. On Gauss10, their utility is nearly identical. On Gauss30, Gauss50 and PAMAP2, \margnet surpasses \aim under sufficient privacy budgets (e.g., $\varepsilon \ge 5.0$). On the Gauss50 dataset, with $\varepsilon=10.0$, \margnet achieved a fidelity error of 0.121, representing a reduction of approximately \textbf{26\%} compared to \aim's error of 0.163.
We attribute this to a fundamental difference: PGM must decompose the joint distribution into local cliques and maintain small clique sizes for computational tractability, forcing it to omit some marginals. In contrast, \margnet can gradually capture all important marginals through adaptive selection without decomposition. We validate this in \Cref{sec: detail comparison}.

{
    \begin{figure*}[!t]
        \centering
        \includegraphics[width=\textwidth]{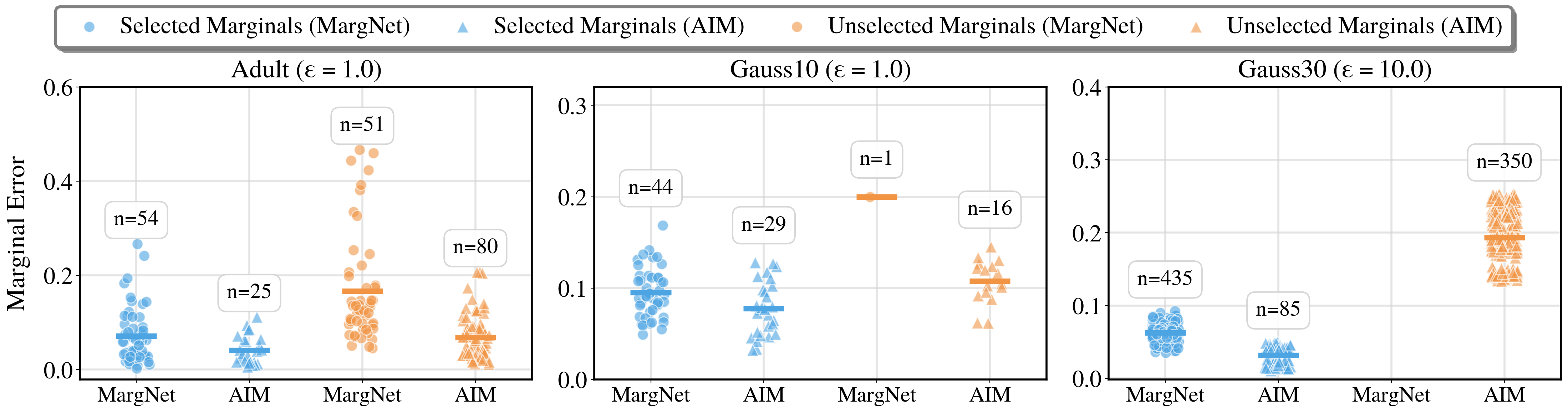}
        \caption{$\ell_1$ distance error of selected and unselected marginals fitting by \aim and \margnet. The three results are obtained on the Adult and Gauss10 datasets under $\varepsilon = 1.0$, and on the Gauss30 dataset under $\varepsilon=10.0$, respectively. Each point in this figure represents a marginal, and the horizontal short lines indicate the mean errors. \rev{If an algorithm does not have marginals in a category, there will be no point presented (e.g., \margnet does not have unselected marginals on Gauss30).}}
        \label{fig: marginal split loss}
    \end{figure*}
}

For \ctgan, it outperforms other methods when $\varepsilon \le 0.2$. We attribute this to PrivTree discretization~\cite{zhang2016privtree} under-splitting under low budgets, causing large discretization errors in other methods while \ctgan outputs continuous results. This can be avoided using discretization methods requiring no privacy budget, such as $k$-uniform binning~\cite{chen2025benchmarkingdifferentiallyprivatetabular} (\Cref{appendix: sup res}). 

An observation that deviates from the findings in \Cref{sec: general comparison} concerns the computational overhead. While \aim continues to suffer from time inefficiency on densely correlated datasets, we observe that the execution time of \margnet increases significantly and consistently as the privacy budget grows, particularly in high-dimensional settings. We attribute this trend to the adaptive nature of our algorithm: as the complexity of the dataset is explicitly recognized during execution, \margnet allocates a correspondingly higher number of computational iterations to measure marginals and optimize the model. Though this process improves the model's performance, it also leads to a higher running time.

\subsection{Detailed Comparison}
\label{sec: detail comparison}

\Cref{sec: general comparison} and \Cref{sec: dense comparison} provided a high-level comparison of utility and time efficiency for multiple algorithms. In this section, we conduct a deep investigation by answering two questions: 
\begin{enumerate}[leftmargin=*, label=\textbullet]
    \item Given an identical set of selected marginals, which model (NN or PGM) achieves higher fidelity when fitting those selected marginals and, crucially, when inferring the unselected marginals?
    \item How does \margnet overcome the weaknesses that \aim encounters in processing datasets that contain densely distributed correlations (high-dimensional cliques) in practice?
\end{enumerate}

\subsubsection{Experimental Settings} \rev{Our experimental settings in this section are listed as follows. We mainly consider three datasets: Adult, Gauss10, and Gauss30 for simplicity, and consider \aim and \margnet. When comparing the fitting ability of PGM and neural networks, we focus on the $\ell_1$ error of the two-way marginals, which serve as the direct inputs to the algorithms, enabling a cleaner comparison of their fitting capabilities.}

\subsubsection{Experimental Results}
In \Cref{fig: random marginal loss}, we initialize PGM and NN with all one-way marginals, and randomly select two-way marginals to fit the model. The number of two-way marginals serves as the x-axis. We record the average $\ell_1$ error of both selected marginals and unselected marginals.
In \Cref{fig: marginal split loss}, we consider three cases: (1) Adult under $\varepsilon=1.0$, where \aim outperform \margnet in synthesis utility; (2) Gauss10 under $\varepsilon=1.0$, where \margnet and \aim have similar performances; (3) Gauss30 under $\varepsilon=10.0$, where \margnet shows superior performance over \aim. We plot the error of each marginal estimated by \margnet and \aim, respectively, to compare the actual fitting error.

\vspace{1.2mm}
\noindent \textbf{Given an identical set of marginals, PGM shows stronger performance in fitting selected marginals, but the estimation error of unselected marginals depends on the dataset's characteristics}. In \Cref{fig: random marginal loss}, PGM shows superior marginal estimation ability on the Adult dataset. Both its selected marginal error and unselected marginal error are lower than those of the neural network. We hypothesize that this is caused by the sparse correlations in the Adult dataset. This sparsity ensures that most unselected marginals can be accurately approximated either by assuming independence or through conditional independence estimations.

However, the situation is different on the Gauss10 and Gauss30 datasets. In these cases, PGM still exhibits more robust estimation of selected marginals, yet it is inferior to the neural network when inferring unselected marginals. We posit that this is due to the overly dense internal correlations within the data, which cause the (conditional) independence assumptions to fail.

\vspace{1.2mm}
\noindent \textbf{In practice, \margnet's ability to select and fit a larger set of marginals enhances its model performance, which in some cases allows it to achieve lower marginal error compared with \aim.} In \Cref{fig: marginal split loss}, we can observe that, on the Adult dataset, even though \margnet selects more marginals, a certain number remain unselected, which corresponds to its larger fidelity error (as shown in \Cref{fig: all combined}). We attribute this to the limited privacy budget, which restricts the number of marginals that can be selected. On the Gauss10 dataset, while \margnet selects most marginals, \aim can also handle many of them due to the dataset's relatively low dimensionality. On Gauss30, however, we observe that \aim can only afford a small proportion of marginals, leaving approximately 80\% unselected. This results in its inferior performance compared to \margnet, which successfully handles all marginals. These results explain the findings from the previous two sections and also reinforce the motivation for this work.
\begin{table}[t]
    \centering
    \caption{\rev{The fidelity error of \aim and \margnet on Gauss-$X$ datasets, where $X$ refers to the dataset dimension.}}
    \label{tab: dim_tvd}
    \resizebox{\columnwidth}{!}{%
    \begin{tabular}{l|ccc|ccc}
    \toprule
    \multirow{2}{*}{Dim} & \multicolumn{3}{c|}{$\epsilon=5.0$} & \multicolumn{3}{c}{$\epsilon=10.0$} \\
    \cmidrule(lr){2-4} \cmidrule(lr){5-7}
     & \aim & \margnet & Abs. Reduction & \aim & \margnet & Abs. Reduction \\
    \midrule
    10 & 0.242 & 0.240 & 0.002 & 0.233 & 0.224 & 0.008 \\
    \rowcolor{gray!20} 30 & 0.181 & 0.158 & 0.023 & 0.161 & 0.132 & 0.029 \\
    50 & 0.187 & 0.153 & 0.033 & 0.163 & 0.121 & 0.042 \\
    \bottomrule
    \end{tabular}%
    }
\end{table}

\begin{table}[t]
    \centering
    \caption{\rev{The fidelity error of \aim and \margnet on Adult and Gauss30 across different $\varepsilon$.}}
    \label{tab: eps_tvd}
    \resizebox{\columnwidth}{!}{%
    \begin{tabular}{l|ccc|ccc}
    \toprule
    \multirow{2}{*}{$\varepsilon$} & \multicolumn{3}{c|}{Adult} & \multicolumn{3}{c}{Gauss30} \\
    \cmidrule(lr){2-4} \cmidrule(lr){5-7}
     & \aim & \margnet & Abs. Reduction & \aim & \margnet & Abs. Reduction \\
    \midrule
    2.0  & 0.024 & 0.045 & -0.021 & 0.209 & 0.198 & 0.011 \\
    \rowcolor{gray!20} 5.0  & 0.018 & 0.024 & -0.006 & 0.181 & 0.158 & 0.023 \\
    10.0 & 0.016 & 0.022 & -0.006 & 0.161 & 0.132 & 0.029 \\
    \bottomrule
    \end{tabular}%
    }
\end{table}

\subsection{\rev{Analysis of Performance Factors}}
\label{sec: exp factor}

\rev{ The preceding experiments suggest that the relative performance of \aim and \margnet depends on multiple dataset and privacy characteristics. In this section, we further examine three of them: data dimensionality, privacy budget, and correlation density. Our goal is not to derive a deterministic algorithm-selection rule, but to investigate how these factors shift the relative strengths of the two methods. In particular, correlation density is a multifaceted property that cannot be completely characterized by a single statistic.
}

\subsubsection{\rev{Experimental Settings}} 
\rev{As this section provides a more in-depth investigation of the preceding experiments, we primarily reuse the previous datasets and metrics used in \Cref{sec: detail comparison}. We focus on the two best algorithms: \aim and \margnet. To characterize pairwise dependence, we use two complementary measures:} 
\begin{enumerate}[leftmargin=*, label=\textbullet]
    \item \rev{\textit{Average absolute Pearson correlation coefficient}. For continuous Gaussian data, we use the average absolute Pearson correlation coefficient. The Pearson correlation coefficient is defined by $\rho(X, Y) = \frac{\text{Cov}(X, Y)}{\sigma_{X} \sigma_{Y}}$ represents the Pearson correlation coefficient between variables $X$ and $Y$. We compute the average absolute coefficient across all pairwise attributes.}
    
    \item \rev{\textit{Average Normalized Mutual Information (NMI)~\cite{JMLR:v11:vinh10a}}. Pearson correlation captures only linear associations and is not generally meaningful for nominal categorical attributes under arbitrary numerical encodings. We therefore use average normalized mutual information (NMI)~\cite{JMLR:v11:vinh10a} as a diagnostic for heterogeneous real-world datasets. For two attributes $X$ and $Y$, we define
    \[ 
        \text{NMI}(X, Y) = \frac{I(X; Y)}{\sqrt{H(X) H(Y)}}, 
    \]
    where $I(X; Y)$ denotes the mutual information between $X$ and $Y$, and $H(\cdot)$ represents the marginal entropy of the distribution. Before computing NMI, numerical attributes are discretized using PrivTree preprocessing adopted in the experiments with $\varepsilon=10$.}
\end{enumerate}
\rev{We must emphasize that neither of these two statistics fully characterize
correlation density. In particular, datasets with the same average
value may differ substantially in how dependencies are distributed
across attribute pairs.}

\begin{figure}[t]
    \centering
    \includegraphics[width=0.98\linewidth]{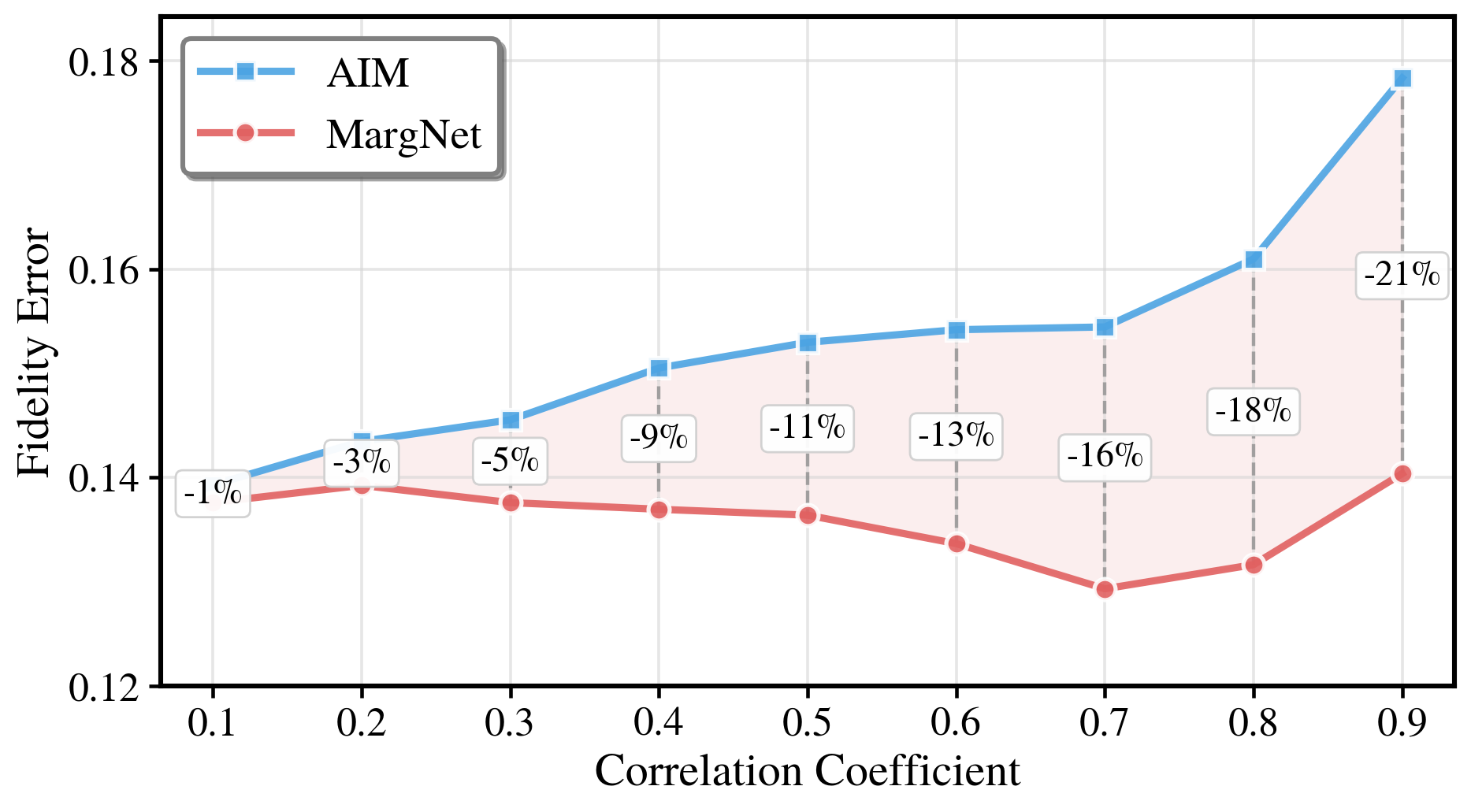}
    \caption{\rev{The fidelity error of \margnet and \aim under $\varepsilon=10$ on Gauss30 built on different correlation coefficients.}}
    \label{fig: gauss_x}
\end{figure}

\subsubsection{\rev{Experimental Results}}
\rev{We investigate three factors: data dimension, privacy budget, and pairwise-dependence structure.}

\vspace{1.2mm}
\noindent\textbf{\rev{Data Dimension}}. \rev{Data dimensionality has previously been identified as an important factor in the relative performance of graphical and neural synthesizers. Our experiments provide complementary evidence under controlled dense-correlation settings. As shown in \Cref{tab: dim_tvd}, when the pairwise correlation structure is fixed and the privacy budget is sufficiently large ($\varepsilon=10$), \margnet begins to outperform \aim at $d=30$. It shows that in the equicorrelated Gaussian setting studied here, dense pairwise dependencies can expose the computational limitations of graphical models at a substantially lower dimensionality.}

\begin{table*}[t]
\centering
\footnotesize
\caption{\rev{Performance variation of AIM and \margnet across multiple independent training runs on Adult ($\varepsilon=1.0$) and Gauss30 ($\varepsilon=10.0$). Each result is aggregated over five independent runs. Std. denotes the standard deviation across runs, and CV is computed as Std./$|\mathrm{Value}|$.}}
\label{tab: stability}
\setlength{\tabcolsep}{4.8pt}
\resizebox{0.88\textwidth}{!}{
\begin{tabular}{ll|cc>{\columncolor{gray!20}}c|cc>{\columncolor{gray!20}}c|cc>{\columncolor{gray!20}}c}
\toprule
& & \multicolumn{3}{c|}{ML Efficacy} & \multicolumn{3}{c|}{Query Error} & \multicolumn{3}{c}{Fidelity Error} \\
\cmidrule(lr){3-5}\cmidrule(lr){6-8}\cmidrule(lr){9-11}
\multicolumn{1}{l}{Dataset} & Method & {   Value   } & {     Std.     } & {   CV   } & {   Value   } & {     Std.     } & {   CV   } & {   Value   } & {     Std.     } & {   CV   } \\
\midrule
\multicolumn{1}{l|}{\multirow{2}{*}{Adult ($\varepsilon=1.0$)}} & AIM & 0.78 & 1.8e-3 & 0.24\% & 0.012 & 4.8e-5 & 0.40\% & 0.03 & 1.4e-3 & 4.38\% \\
\multicolumn{1}{l|}{} & {\margnet \quad} & 0.76 & 6.9e-3 & 0.92\% & 0.013 & 1.2e-4 & 0.93\% & 0.07 & 4.6e-3 & 6.46\% \\
\midrule
\multicolumn{1}{l|}{\multirow{2}{*}{Gauss30 ($\varepsilon=10.0$) \quad}} & AIM & -- & -- & \cellcolor{white}-- & 0.006 & 1.0e-4 & 1.75\% & 0.16 & 2.1e-3 & 1.27\% \\
\multicolumn{1}{l|}{} & \margnet & -- & -- & \cellcolor{white}-- & 0.005 & 1.2e-4 & 2.38\% & 0.13 & 4.8e-4 & 0.36\% \\
\bottomrule
\end{tabular}%
}
\end{table*}

\vspace{1.2mm}
\noindent\textbf{\rev{Privacy Budget}}. \rev{The privacy budget also changes the relative strengths of the two methods. Under a limited budget, both methods can measure only a small collection of informative marginals, while \aim can further infer unmeasured distributions through conditional-independence
assumptions. As the budget increases, \margnet can select and fit more marginals directly. In our Gauss30 experiments, \margnet begins to outperform \aim at $\varepsilon=2$, and the advantage becomes more pronounced for $\varepsilon \geq 5$. These values describe the trend observed in this particular setting rather than a general privacy-budget threshold.}

\vspace{1.2mm}
\noindent\textbf{\rev{Pairwise-Dependence Structure}}. \rev{Finally, we investigate the influence of correlation density. We first use jointly Gaussian data, whose dependence structure is characterized by the covariance matrix. This allows us to vary the common pairwise Pearson correlation coefficient while fixing the dimensionality, privacy budget, and marginal distributions, thereby isolating the effect of pairwise dependence strength. As shown in \Cref{fig: gauss_x}, the relative advantage of \margnet generally increases as the common pairwise correlation becomes stronger. A noticeable performance gap begins to emerge around a coefficient of $0.3$ in this controlled Gauss30 setting. This result supports our hypothesis that broadly distributed strong
dependencies make conditional-independence approximations less effective and require graphical models to maintain increasingly complex structures. However, the value $0.3$ should be interpreted only as an observation from this controlled experiment, rather than as a general threshold for distinguishing the two methods.}

\rev{Pearson correlation provides a useful control variable in our Gaussian experiments, but it is insufficient for characterizing correlation density in heterogeneous real-world datasets. It captures only linear associations and is not directly applicable to nominal categorical attributes without imposing an arbitrary numerical encoding. For example, Loan and PAMAP2 have similar average absolute Pearson coefficients over their numerical attributes, $0.12$ and
$0.13$, respectively, while their average NMI values are $0.03$ and $0.09$. This discrepancy illustrates that a single linear summary may overlook nonlinear or categorical dependencies present in real data. Nevertheless, NMI does not provide a complete characterization either: its value depends on the discretization procedure, and an average score may obscure how dependencies are distributed across different attribute pairs. Therefore, Pearson correlation and NMI should be regarded as complementary descriptive indicators rather
than universal criteria for distinguishing correlation regimes or selecting between \aim and \margnet.}

\subsection{\rev{Variation Analysis}}

\rev{Finally, we evaluate the performance variation of \margnet and \aim across independent runs of the complete synthesis pipeline.}

\subsubsection{\rev{Experimental Settings}}
\rev{We independently run \margnet and \aim five times on two representative settings: Adult with $\varepsilon=1.0$ (sparsely correlated datasets under a limited privacy budget) and Gauss30 with $\varepsilon=10.0$ (densely correlated datasets under a sufficient privacy budget). We report the mean, standard deviation, and coefficient of variation (CV) across the five runs.}

\subsubsection{\rev{Experimental Results}}
\rev{As shown in \Cref{tab: stability}, both methods exhibit limited variation. On Adult, \margnet shows somewhat higher variation than \aim, likely due to the additional randomness of neural optimization and the limited privacy budget, under which some dependencies remain insufficiently constrained. Both methods remain stable on Gauss30. More importantly, the performance gaps are substantially larger than the corresponding standard deviations: \aim maintains its advantage on Adult, whereas \margnet consistently achieves lower errors on Gauss30. Therefore, run-to-run randomness does not change the qualitative conclusions in these representative settings.}

\section{Conclusions}
\label{sec: conclusion}

In this work, we revisited the comparison between graphical and neural approaches to DP tabular data synthesis through the lens of data correlation regimes. We proposed MargNet, a feature-DP neural synthesis framework that integrates adaptive full marginal selection into differentiable NN training, and evaluated it on both sparsely correlated and densely correlated datasets. Our results show that the relative strengths of different synthesis paradigms depend strongly on dataset characteristics. On sparsely correlated datasets, AIM remains a very strong baseline and often achieves the highest utility, while MargNet achieves the best synthesis utility among the compared methods on the densely correlated datasets we evaluate. Taken together, these findings suggest that NN-based methods remain an important direction for DP tabular synthesis and that algorithm selection should account for the data regime rather than follow a one-size-fits-all view.

Our study also has several limitations. \rev{The evidence for densely correlated regimes is based on a limited number of datasets and is largely empirical, so the observed pattern should not be viewed as a universal rule.} Although we analyze MargNet's fitting behavior, we do not formally characterize when neural methods outperform graphical models. Moreover, correlation is only one factor affecting synthesis difficulty; dimensionality, domain size, and distribution shape may also matter. Future work should develop more principled measures of data regimes, evaluate broader benchmarks, and improve how private neural models incorporate distributional information.

\section*{Acknowledgements}

This work is supported by NSF CNS-2220433, OAC-2319988, CNS-2543284, and CCI (Virginia Commonwealth Cyber Initiative)

\bibliographystyle{ACM-Reference-Format}
\balance
\bibliography{ref}

\ifnum\value{includeSecuritySection}=1

\appendix

\section*{Generative AI Usage}

LLMs were used for editorial purposes in this manuscript, and all outputs were inspected by the authors to ensure accuracy and originality. LLM-assisted editing was mainly applied to \Cref{sec: intro,sec: exist work,sec: margnet,sec: exp,sec: conclusion}. LLMs were not used in methodology development.

\section*{Ethical Considerations}


This work focuses on DP tabular data synthesis, with the goal of enabling useful data analysis and generation while protecting the privacy of individuals in the original dataset. Our research is primarily defensive in nature: the proposed methods are designed to reduce privacy risks rather than facilitate attacks or unauthorized access to sensitive information. Therefore, we do not identify significant additional ethical risks arising from the methodology itself beyond the general limitations of DP and synthetic data.

\section*{Open Science}

All code, execution instructions, and related datasets are provided in our GitHub repository. 

\fi

%% file: appendix.tex
\clearpage

\appendix

\section{Data Preprocessing}

Aligning with previous works~\cite{chen2025benchmarkingdifferentiallyprivatetabular, zhang2021privsyn, mckenna2022aim, tao2022benchmarkingdifferentiallyprivatesynthetic}, we also need to utilize some preprocessing methods to make sure all methods can work across different datasets. 

\begin{enumerate}[leftmargin=*, label=\textbullet]
    \item \textbf{PrivTree Discretization}. PrivTree algorithm~\cite{zhang2016privtree} can divide a continuous value range under differential privacy. Here, we apply PrivTree to \aim, \margnet, \gem, \privsyn, and \rapp in Gaussian datasets. 

    \item \textbf{Rare Category Filtering}. We follow the filtering method for categorical attributes proposed by Chen et al.~\cite{chen2025benchmarkingdifferentiallyprivatetabular}, which introduces both a $3\sigma$ criterion~\cite{mckenna2021winning, zhang2021privsyn} and a fixed filter threshold, guaranteeing the simplicity of the dataset.
\end{enumerate}

\section{Detailed Proof}
\label{appendix: proof}

\subsection{Proof of \Cref{theorem: q}}

\begin{proof}
\rev{Because $n_i$ and $\rho_m$ are irrelevant to the data, we have 
\[
\Delta_q = r_i\Delta\left(\left\lVert M_i\left(G^{k-1}\right) - M_i\right\rVert_1\right)
\]
For any $D$ and $D'$, because $M_i(D)$ and $M_i(D')$ only differ in one entry with a gap of 1 ($\left|M_i(D)[j] - M_i(D')[j]\right| = 1$), we can obtain 
\[
\begin{aligned}
& \left\lVert \lVert M_i\left(G^{k-1}\right) - M_i(D)\rVert_1 - \lVert M_i\left(G^{k-1}\right) - M_i(D')\rVert_1 \right\rVert_2 \\
\leq & \lVert [0,\cdots,1,\cdots,0]^T \lVert_2 = 1
\end{aligned}
\]
Therefore, $\Delta_q = r_i$.}
\end{proof}

\subsection{Proof of \Cref{theorem: privacy}}

\begin{proof} \rev{The initialization step measures $d$ one-way marginals, each satisfying $\rho_m$-zCDP. By composition, this step satisfies
$d\rho_m$-zCDP, leaving $\rho_{\mathrm{total}}=\rho-d\rho_m$ for \Cref{algo: adapt}. In each iteration of \Cref{algo: adapt}, the exponential-mechanism selection and Gaussian measurement satisfy $\rho_s$-zCDP and $\rho_m$-zCDP, respectively. All subsequent model training and data generation are post-processing. Moreover, whenever the budget proposed for the next iteration reaches the remaining budget, the algorithm sets
\[
\rho_s=0.1(\rho_{\mathrm{total}}-\rho_{\mathrm{used}}),
\quad
\rho_m=0.9(\rho_{\mathrm{total}}-\rho_{\mathrm{used}}).
\]
Thus, the cumulative privacy cost of \Cref{algo: adapt} never exceeds $\rho_{\mathrm{total}}$. By adaptive composition, \Cref{algo: adapt} satisfies $\rho_{\mathrm{total}}$-zCDP. Therefore, the total privacy cost of Algorithm~1 is at most $d\rho_m+\rho_{\mathrm{total}}=\rho$, which completes the proof.}
\end{proof}

\subsection{Proof of \Cref{theorem: selected lower bound}}

\begin{proof} First, we present a result from Eckart and Young~\cite{eckart1936approximation}.
\begin{lemma}\label{lemma: matrix fit}
    Let $\lambda_1, \ldots, \lambda_r$ be the singular values of $P$, sorted in descending order. Then the minimum approximation error when approximating $P$ by a rank-$m$ matrix is 
    \[
    \min_{\substack{Q \\ \operatorname{rank}(Q) \leq m}} \left\lVert P-Q\right\rVert_F^2 = \sum_{t=m+1}^r \lambda_t^2.
    \]
\end{lemma}

For a batch of samples, we use the average of the distribution of each row as the final marginal estimation. Therefore, the error $\mathcal{L}(G)$ can be written as
\[
\mathcal{L}_s(G) = \sum_{i \in I_s}\left\lVert M_i - \frac{\hat{N}}{b} \sum_{j=1}^{b} u^T_j v_j \right\rVert_F^2,
\]
where $u_j$ and $v_j$ are the one-way marginals of two attributes in the $j$-th sample row. (If $M_i$ itself is a one-way marginal, we can simply set $v_j$ to be a $1\times1$ matrix.) Note that the rank of $\sum_{j=1}^b u^T_j v_j$ is at most $b$ since it is a sum of $b$ rank-one matrices. By \Cref{lemma: matrix fit}, we have 
\[
     \mathcal{L}_s(G) = \sum_{i \in I_s}\left\lVert \frac{1}{b}\sum_{j=1}^b u^T_j v_j - M_i\right\rVert_F^2 \geq \sum_{i \in I_s} \left(\sum_{t=b+1}^{r(M_i)} \lambda_t^2\right),
\]
which is a consequence of \Cref{theorem: selected lower bound}.
\end{proof}

\subsection{Proof of \Cref{theorem: selected upper bound}}

\begin{proof} We first decompose $\mathcal{L}_s(G)$ into two terms. By the inequality $\|A+B\|_F^2 \leq 2(\|A\|_F^2 + \|B\|_F^2)$, we have
\begin{equation} \label{eq: upper bound comp}
\begin{aligned}
    \mathcal{L}_s(G) & = \sum_{i \in I_s}\left\lVert M_i - \bar{M}_i + \bar{M}_i - \hat{M}_i(G)\right\rVert_F^2\\
    & \leq 2\sum_{i \in I_s}\left(\left\lVert \bar{M}_i - \hat{M}_i(G)\right\rVert_F^2 + \left\lVert M_i - \bar{M}_i \right\rVert_F^2 \right).
    \end{aligned}
\end{equation}
Note that $\lVert \bar{M}_i - \hat{M}_i(G)\rVert_F^2$ is computable from our model, so it suffices to bound $\lVert M_i - \bar{M}_i \rVert_F^2$. Recall that $\bar{M}_i = M_i + \mathcal{N}(0, \bar{\sigma}^2_i\mathbb{I})$, which implies
\[
    M_i - \bar{M}_i \sim \mathcal{N}(0, \bar{\sigma}^2_i)^{n_i}.
\]
Consequently,
\[
    \left\lVert M_i - \bar{M}_i \right\rVert_F^2 \sim \bar{\sigma}^2_i \chi_{n_i}^2,
\]
where $\chi_{n_i}^2$ denotes the chi-squared distribution with $n_i$ degrees of freedom. Let $F^{-1}_{\chi_{n_i}^2}$ denote the inverse cumulative distribution function of $\chi_{n_i}^2$. Then for any $\delta_i \in (0,1)$, with probability at least $1-\delta_i$,
\[
\left\lVert M_i - \bar{M}_i \right\rVert_F^2 \leq \bar{\sigma}_i^2 F^{-1}_{\chi_{n_i}^2}(1-\delta_i).
\]
Applying this bound to \Cref{eq: upper bound comp} yields \Cref{theorem: selected upper bound}.
\end{proof}

{
\begin{table*}
    \centering 
    \caption{Results of \merf and \ddpm on sparsely correlated datasets.}
    \label{tab: sup general}
    \resizebox{\textwidth}{!}{
    \begin{tabular}{l|l|rrrrrr|rrrrrr|rrrrrr}
    \toprule
     \multirow{2}{*}{Dataset} & \multirow{2}{*}{Method} & \multicolumn{6}{c|}{ML Efficacy} & \multicolumn{6}{c|}{Query Error} & \multicolumn{6}{c}{Fidelity Error} \\
    \cmidrule{3-20}
     &   & 0.2 & 0.5 & 1.0 & 2.0 & 5.0 & 10.0 & 0.2 & 0.5 & 1.0 & 2.0 & 5.0 & 10.0 & 0.2 & 0.5 & 1.0 & 2.0 & 5.0 & 10.0 \\
    \midrule
    \multirow{2}{*}{Adult} & DP-MERF & 0.63 & 0.73 & 0.71 & 0.72 & 0.72 & 0.72 & 0.019 & 0.018 & 0.018 & 0.018 & 0.017 & 0.016 & 0.324 & 0.296 & 0.265 & 0.266 & 0.218 & 0.267 \\
     & DP-TabDDPM & 0.43 & 0.43 & 0.43 & 0.43 & 0.43 & 0.43 & 0.075 & 0.045 & 0.039 & 0.038 & 0.039 & 0.038 & 0.889 & 0.443 & 0.378 & 0.341 & 0.328 & 0.321 \\
    \midrule
    \multirow{2}{*}{Bank} & DP-MERF & 0.61 & 0.61 & 0.63 & 0.62 & 0.61 & 0.61 & 0.012 & 0.011 & 0.008 & 0.009 & 0.009 & 0.010 & 0.332 & 0.312 & 0.274 & 0.280 & 0.280 & 0.299 \\
     & DP-TabDDPM & 0.47 & 0.47 & 0.47 & 0.47 & 0.47 & 0.47 & 0.105 & 0.092 & 0.070 & 0.068 & 0.057 & 0.053 & 0.474 & 0.649 & 0.754 & 0.761 & 0.715 & 0.703 \\
    \midrule
    \multirow{2}{*}{Loan} & DP-MERF & 0.22 & 0.18 & 0.24 & 0.21 & 0.21 & 0.18 & 0.058 & 0.063 & 0.063 & 0.060 & 0.060 & 0.059 & 0.902 & 0.894 & 0.887 & 0.904 & 0.922 & 0.904 \\
     & DP-TabDDPM & 0.24 & 0.24 & 0.24 & 0.25 & 0.26 & 0.25 & 0.059 & 0.055 & 0.059 & 0.055 & 0.054 & 0.056 & 0.901 & 0.896 & 0.891 & 0.885 & 0.873 & 0.874 \\
    \bottomrule
    \end{tabular}
    }
\end{table*}
}

\subsection{Proof of \Cref{theorem: final unselected bound}}

\begin{proof} To prove \Cref{theorem: final unselected bound}, we first give a primary conclusion~\cite{dwork2014algorithmic}. 

\begin{lemma} \label{lemma: exp prop}
Assuming the final selected marginal during the adaptive framework is $M_K$, then with at least probability $1-\delta$, the following inequality holds for any $i \ne K$.
\begin{equation} \label{eq: exp prop}
    q(i) \leq q(K) + \frac{\Delta_q}{\sqrt{2 \rho_s^k}}\log(|C|/\delta),
\end{equation}
where $C$ is the candidate set, $q_i$ is defined in \cref{algo: adapt}.
\end{lemma}

For the final model $G$, we can build the error estimation with the model before the last selection step $G^{K-1}$. Assuming that we have conducted $K$ selection steps and marginal $M_i$ is not selected, and the marginal fitted by the model $G$ and intermediate model $G^{K-1}$ are $\hat{M}_i$ and $\hat{M}^{K-1}_i$, respectively, we have 
\begin{equation} \label{eq: unselected bound}
    \left\lVert M_i - \hat{M}_i\right\rVert_1 \;\leq\; \left\lVert M_i - \hat{M}^{K-1}_i\right\rVert_1 + \left\lVert  \hat{M}^{K-1}_i - \hat{M}_i\right\rVert_1
\end{equation}
Here, we bound the fitting error of the unselected marginal by two terms: one is the fitting error in the last selection round, and the other is the improvement from the model fitting after the last selection round.

Then, by \Cref{lemma: exp prop}, we have that with probability $1-\delta$, 
\begin{equation*}
\begin{aligned}
r_i\Bigl( & \left\lVert M_i - \hat{M}^{K-1}_i\right\rVert_1 - n_i/\sqrt{\pi \rho_K^S}\Bigr) \\ 
& \leq r_\theta \Bigl( \left\lVert M_\theta - \hat{M}^{K-1}_\theta \right\rVert_1 - n_\theta/\sqrt{\pi \rho_K^S}\Bigr) + \frac{\Delta_q}{\sqrt{2 \rho_s^k}}\log\frac{|C|}{\delta}
\end{aligned}
\end{equation*}

Rearranging the above inequality, we obtain that for any unselected marginal $M_i$, and the finally selected marginal $M_\theta$,
\begin{equation}
\begin{aligned}
    \left\lVert M_i - \hat{M}^{K-1}_i\right\rVert_1 & \leq \frac{r_\theta}{r_i} \left\lVert M_\theta - \hat{M}^{K-1}_\theta\right\rVert_1 \\
    & + \frac{n_ir_i - n_\theta r_\theta}{r_i\sqrt{\pi \rho_s^K}} + \frac{\Delta_q}{r_i\sqrt{2 \rho_s^K}}\log\frac{|C|}{\delta}
\end{aligned}
\end{equation}
holds with at least probability $1-\delta$. $r_i$ and $r_\theta$ are the weights used in exponential mechanism in \cref{algo: adapt}, and $\rho_s^K$ is the value of $\rho_s$ in the last selection iteration.

We denote the RHS of \Cref{eq: exp bound} as $B_{i, K}$. Then by \Cref{eq: unselected bound}, we can obtain an upper bound of the total $\ell_1$ error of unselected marginals. Then with \Cref{eq: unselected bound}, we have 
\begin{equation*}
    \sum_i \left\lVert M_i - \hat{M}_i\right\rVert_1 \;\leq\; \sum_i B_{i, K} + \sum_i \left\lVert  \hat{M}^{K-1}_i - \hat{M}_i\right\rVert_1 ,
\end{equation*}
as claimed in \Cref{theorem: final unselected bound}.
\end{proof}

\begin{table}[t]
\footnotesize
\centering
\caption{Hyperparameters of \margnet (sparsely correlated datasets). $T$ and $T^*$ are training iterations under $0.2\leq \varepsilon \leq 5.0$ and $\varepsilon = 10.0$, respectively.}
\label{tab: hyper_part1}
\resizebox{\columnwidth}{!}{
\begin{tabularx}{\columnwidth}{l|XXX}
     \toprule
     & Adult & Bank & Loan \\
     \midrule
     Training Iteration $T$ & $200$ & $200$ & $1000$ \\
     Training Iteration $T^*$ & $100$ & $100$ & $1000$ \\
     \midrule
     Learning Rate $r$ & $0.001$ & $0.001$ & $0.001$ \\
     Batch Size $b$ & $512$ & $512$ & $512$ \\
     Parameter $c$ & $16d$ & $16d$ & $16d$ \\
     Parameter $r_i$ & $1.0$ & $1.0$ & $1.0$ \\
     \bottomrule
\end{tabularx}
}
\end{table}

\begin{table}[t]
\footnotesize
\centering
\caption{Hyperparameters of \margnet (densely correlated datasets). $T$ and $T^*$ are training iterations under $0.2\leq \varepsilon \leq 5.0$ and $\varepsilon = 10.0$, respectively. Here G$\cdot$ is the abbreviation of Gauss$\cdot$.}
\label{tab: hyper_part2}
\resizebox{\columnwidth}{!}{
\begin{tabularx}{\columnwidth}{l|XXXX}
     \toprule
     & Gauss10 & Gauss30 & Gauss50 & PAMAP2 \\
     \midrule
     Training Iteration $T$ & $200$ & $200$ & $500$ & $200$ \\
     Training Iteration $T^*$ & $100$ & $100$ & $500$ & $200$ \\
     \midrule
     Learning Rate $r$ & $0.001$ & $0.001$ & $0.001$ & $0.001$ \\
     Batch Size $b$ & $256$ & $1024$ & $1024$ & $2048$ \\
     Parameter $c$ & $16d$ & $16d$ & $16d$ & $16d$ \\
     Parameter $r_i$ & $1.0$ & $1.0$ & $1.0$ & $1.0$ \\
     \bottomrule
\end{tabularx}
}
\end{table}

{
\begin{table*}
    \centering 
    \caption{Results of \merf, \ddpm, \privsyn on densely correlated datasets. Note that \privsyn encounters an Out-of-Memory error on the PAMAP2 dataset, so those results are omitted.}
    \label{tab: sup gauss}
    \resizebox{\textwidth}{!}{
    \begin{tabular}{l|l|rrrrrr|rrrrrr|rrrrrr}
    \toprule
     \multirow{2}{*}{Dataset} & \multirow{2}{*}{Method} & \multicolumn{6}{c|}{{Query Error}} & \multicolumn{6}{c|}{{Fidelity Error}} & \multicolumn{6}{c}{{Running Time (min)}}\\
    \cmidrule{3-20}
     &   & 0.2 & 0.5 & 1.0 & 2.0 & 5.0 & 10.0 & 0.2 & 0.5 & 1.0 & 2.0 & 5.0 & 10.0 & 0.2 & 0.5 & 1.0 & 2.0 & 5.0 & 10.0 \\
    \midrule
    \multirow{3}{*}{Gauss10} & DP-MERF & 0.223 & 0.037 & 0.192 & 0.027 & 0.028 & 0.027 & 0.985 & 0.475 & 0.966 & 0.456 & 0.478 & 0.420 & 0.06 & 0.06 & 0.20 & 0.06 & 0.16 & 0.06 \\
     & DP-TabDDPM & 0.216 & 0.195 & 0.201 & 0.196 & 0.200 & 0.194 & 0.906 & 0.855 & 0.877 & 0.835 & 0.879 & 0.862 & 0.87 & 0.67 & 0.76 & 0.76 & 0.76 & 0.76 \\
     & PrivSyn & 0.133 & 0.140 & 0.101 & 0.071 & 0.065 & 0.060 & 0.801 & 0.819 & 0.716 & 0.659 & 0.653 & 0.639 & 0.26 & 0.34 & 0.29 & 0.28 & 0.47 & 0.32 \\
    \midrule
    \multirow{3}{*}{Gauss30} & DP-MERF & 0.217 & 0.225 & 0.209 & 0.233 & 0.049 & 0.049 & 0.963 & 0.969 & 0.947 & 0.956 & 0.347 & 0.392 & 0.12 & 0.16 & 0.12 & 0.12 & 0.12 & 0.12 \\
     & DP-TabDDPM & 0.234 & 0.213 & 0.199 & 0.207 & 0.202 & 0.196 & 0.955 & 0.907 & 0.879 & 0.893 & 0.880 & 0.865 & 8.95 & 8.77 & 9.08 & 8.83 & 8.68 & 8.86 \\
     & PrivSyn & 0.149 & 0.126 & 0.103 & 0.083 & 0.079 & 0.064 & 0.824 & 0.774 & 0.707 & 0.661 & 0.667 & 0.625 & 3.16 & 2.38 & 3.43 & 3.76 & 3.69 & 4.55 \\
    \midrule
    \multirow{3}{*}{Gauss50} & DP-MERF & 0.018 & 0.016 & 0.018 & 0.032 & 0.019 & 0.035 & 0.274 & 0.297 & 0.335 & 0.307 & 0.287 & 0.338 & 0.20 & 0.17 & 0.22 & 0.18 & 0.17 & 0.17 \\
     & DP-TabDDPM & 0.239 & 0.251 & 0.242 & 0.248 & 0.245 & 0.249 & 0.959 & 0.983 & 0.965 & 0.979 & 0.972 & 0.978 & 23.09 & 22.54 & 22.39 & 16.59 & 15.54 & 15.52 \\
     & PrivSyn & 0.148 & 0.142 & 0.109 & 0.089 & 0.066 & 0.060 & 0.821 & 0.809 & 0.725 & 0.675 & 0.625 & 0.613 & 26.53 & 23.72 & 11.02 & 22.70 & 22.47 & 24.33 \\
    \midrule
    \multirow{3}{*}{PAMAP2} & DP-MERF & 0.042 & 0.035 & 0.029 & 0.106 & 0.048 & 0.049 & 0.574 & 0.555 & 0.496 & 0.709 & 0.583 & 0.558 & 0.68 & 0.29 & 0.27 & 0.28 & 0.66 & 0.27 \\
     & DP-TabDDPM & 0.193 & 0.167 & 0.174 & 0.174 & 0.173 & 0.158 & 0.748 & 0.671 & 0.761 & 0.778 & 0.763 & 0.650 & 34.23 & 35.64 & 32.93 & 35.97 & 39.16 & 38.73 \\
     & PrivSyn & - & - & - & - & - & - & - & - & - & - & - & - & 0.00 & - & - & - & - & - \\
    \bottomrule
    \end{tabular}
    }
\end{table*}
}

\section{Implementation Details}
\label{appendix: set}

As introduced earlier, \margnet involves multiple training hyperparameters. We provide the relevant implementation details. The hyperparameter settings are summarized in \Cref{tab: hyper_part1,tab: hyper_part2}. Some extra details are provided as follows. 

\begin{enumerate}[leftmargin=*, label=\textbullet]
    \item \textbf{Network Architecture.} We adopt the same network structure as in prior work~\cite{fang2022dp, liu2021iterative}. The detailed architecture is available in our open-source repository and is omitted here for brevity.
    
    \item \textbf{Parameter $c$.} Following \aim, we set $c=16d$ for all datasets, where $d$ is the dimension of the dataset. This constraint ensures that at most $16d$ marginals are selected during \cref{algo: adapt}, which also motivates our choice of $d$ as the weight enhancement factor for newly selected marginals.
    
    \item \textbf{Training Iterations.} Notably, \textbf{we use fewer training iterations for larger privacy budgets $\varepsilon$} on most datasets. This counterintuitive choice is motivated by the fact that a larger $\varepsilon$ allows selecting more marginals, which in turn increases the effective number of training iterations in our model. Reducing the nominal iteration count in these cases provides two benefits: (1) it prevents overfitting and yields more stable results, and (2) it accelerates model fitting without significantly compromising utility.
\end{enumerate}

\rev{Furthermore, we clarify that the evaluated algorithms possess distinct model architectures and parameters. Therefore, we provide specific configurations for them, which mainly follow the recommendations provided in the existing literature. For instance, the graphical model underlying \aim requires fewer parameters. Certain settings, such as the number of optimization steps, are adaptive (i.e., the total number of optimization iterations equals the number of cliques multiplied by a predefined maximum limit). Consequently, it is sufficient to set an adequately large value for the optimization steps per clique. In our experiments, we run all algorithms under $\delta=1e-5$.}

{
\begin{table}[t]
\footnotesize
\centering
\caption{Comparison of \margnet (with uniform binning) with \ctgan and \merf under $\varepsilon=0.2$.}
\label{tab: uniform res}
\begin{tabularx}{\columnwidth}{X|X|XXX}
     \toprule
     Metric & Dataset& \margnet& \ctgan & \merf\\
     \midrule
     \multirow{3}{*}{Query Err.} 
     & Gauss10 & \textbf{0.025}& 0.055& 0.223\\
     & Gauss30 & \textbf{0.017}& 0.058& 0.217 \\
     & Gauss50 & \textbf{0.017}& 0.035& 0.018\\
     & PAMAP2 & \textbf{0.025}& 0.025& 0.042\\
     \midrule
      \multirow{3}{*}{Fidelity Err.} 
      & Gauss10 & \textbf{0.35}& 0.58& 0.99\\
      & Gauss30 & \textbf{0.26}& 0.57& 0.96 \\
      & Gauss50 & \textbf{0.25}& 0.45& 0.27\\
      & PAMAP2 & \textbf{0.54}& 0.46& 0.57\\
     \bottomrule
\end{tabularx}
\end{table}
}

\section{Supplementary Experimental Results} 
\label{appendix: sup res}

\subsection{Supplementary Results of Baselines}

In \Cref{sec: general comparison} and \Cref{sec: dense comparison}, we discussed \ddpm, \merf, and \privsyn, but do not present their experimental results in all figures and tables. It is because that these methods' performance is inferior to that of other methods. We now present these results here. For \ddpm and \merf's results on real-world datasets, we present them in \Cref{tab: sup general}. The densely correlated dataset results are provided in \Cref{tab: sup gauss}.

On those densely correlated datasets, \merf exhibits results similar to those of \ctgan, achieving strong performance at very low values of $\varepsilon$ on densely correlated datasets. We previously attributed this to the coarse discretization used by marginal-based methods rather than to the strong synthesis ability of \merf or \ctgan. To support this explanation, we present \margnet's performance on densely correlated datasets under uniform binning (bin number $= 10$ for the Gaussian dataset and $20$ for the PAMAP2 dataset) in \Cref{tab: uniform res}. We can easily observe that under uniform binning, \margnet achieves higher utility than \merf and \ctgan in most cases. In the experiments described in the main text, we select PrivTree because it adaptively partitions domains according to the privacy budget, resulting in more precise discretization when the budget is sufficient.

{
\begin{table}[t]
\footnotesize
    \centering
    \caption{Algorithm module ablation study results on the Adult and Gauss10 dataset. ML eff., query err., and fidelity err. are the abbreviations of machine learning efficacy, query error, and fidelity error, respectively.}
    \label{tab: ablation}
    \resizebox{\columnwidth}{!}{
    \begin{tabular}{c|c|c|ccc}
    \toprule
    Dataset & $\varepsilon$ & Method & ML Eff. $\uparrow$ & Query Err. $\downarrow$ & Fidelity Err. $\downarrow$ \\
    \midrule
    \multirow{9}{*}{Adult} & \multirow{3}{*}{0.2} 
       & MargNet-q & 0.66 & 0.016 & 0.17 \\
     & & MargNet-s & 0.73 & 0.015 & 0.16 \\
     \rowcolor{gray!20} \cellcolor{white} & \cellcolor{white} & MargNet & 0.74 & 0.015 & 0.11 \\
    \cmidrule{2-6}
    & \multirow{3}{*}{1.0} 
     & MargNet-q & 0.63 & 0.013 & 0.11 \\
     & & MargNet-s & 0.76 & 0.012 & 0.08 \\
     \rowcolor{gray!20} \cellcolor{white} & \cellcolor{white} & MargNet & 0.76 & 0.012 & 0.06 \\
    \cmidrule{2-6}
    & \multirow{3}{*}{10.0} 
     & MargNet-q & 0.68 & 0.013 & 0.11 \\
     & & MargNet-s & 0.77 & 0.012 & 0.02 \\
     \rowcolor{gray!20} \cellcolor{white} & \cellcolor{white} & MargNet & 0.77 & 0.012 & 0.02 \\
    \midrule
    \multirow{9}{*}{Gauss10} & \multirow{3}{*}{0.2}
        & MargNet-q & - & 0.076 & 0.59 \\
     &  & MargNet-s & - & 0.077 & 0.58 \\
     \rowcolor{gray!20} \cellcolor{white} & \cellcolor{white} & MargNet & - & 0.077 & 0.57 \\
     \cmidrule{2-6}
     & \multirow{3}{*}{1.0} 
        & MargNet-q & - & 0.022 & 0.40 \\
     &  & MargNet-s & - & 0.021 & 0.33 \\
     \rowcolor{gray!20} \cellcolor{white} & \cellcolor{white} & MargNet & - & 0.020 & 0.32 \\
     \cmidrule{2-6}
     & \multirow{3}{*}{10.0} 
        & MargNet-q & - & 0.017 & 0.44 \\
     &  & MargNet-s & - & 0.006 & 0.23 \\
     \rowcolor{gray!20} \cellcolor{white} & \cellcolor{white} & MargNet & - & 0.005 & 0.22 \\
    \bottomrule
    \end{tabular}
    }
\end{table}
}

\begin{table*}[t]
    \centering
    \footnotesize
    \caption{\rev{Four- and five-way query errors of AIM and \margnet on Adult and PAMAP2.}}
    \label{tab:high_way_query}
    \setlength{\tabcolsep}{8pt}
    \begin{tabular}{l|l|rrrrrr|rrrrrr}
    \toprule
    \multirow{2}{*}{Dataset} & \multirow{2}{*}{Method} & \multicolumn{6}{c|}{4-way query error} & \multicolumn{6}{c}{5-way query error} \\
    \cmidrule(lr){3-8}\cmidrule(lr){9-14}
    & & 0.2 & 0.5 & 1.0 & 2.0 & 5.0 & 10.0 & 0.2 & 0.5 & 1.0 & 2.0 & 5.0 & 10.0 \\
    \midrule
    \multirow{2}{*}{Adult} & AIM & 1.4e-2 & 1.3e-2 & 1.4e-2 & 1.3e-2 & 1.3e-2 & 1.2e-2 & 9.4e-3 & 9.0e-3 & 9.2e-3 & 9.0e-3 & 8.7e-3 & 8.5e-3 \\
    & \margnet & 1.5e-2 & 1.4e-2 & 1.4e-2 & 1.3e-2 & 1.3e-2 & 1.3e-2 & 1.1e-2 & 1.0e-2 & 8.8e-3 & 8.4e-3 & 9.5e-3 & 8.9e-3 \\
    \midrule
    \multirow{2}{*}{PAMAP2} & AIM & 2.6e-2 & 1.3e-2 & 8.1e-3 & 4.7e-3 & 3.4e-3 & 2.5e-3 & 2.1e-2 & 1.1e-2 & 6.6e-3 & 3.9e-3 & 2.8e-3 & 2.2e-3 \\
    & \margnet & 2.6e-2 & 1.4e-2 & 8.4e-3 & 4.8e-3 & 3.3e-3 & 2.2e-3 & 2.2e-2 & 1.2e-2 & 6.8e-3 & 3.9e-3 & 2.6e-3 & 1.9e-3 \\
    \bottomrule
    \end{tabular}
\end{table*}

\subsection{Ablation Study of Algorithm Modules} \label{sec: ablation}

To identify which designs contribute to \margnet's superior synthesis utility over prior NN-based methods, we conduct an ablation study in this section. We systematically remove or modify key components of \margnet and compare the performances of these new algorithms with those of \margnet. The metrics we use in this section are kept the same as \Cref{sec: general comparison}. Here, we introduce the datasets and baselines.
 
\noindent \textbf{Datasets}. To keep the experiments simple yet representative, we conduct ablation studies on the Adult and Gauss10 datasets, which respectively represent the sparsely and densely correlated datasets. We set the privacy budget to $\varepsilon \in \{0.2, 1.0, 10.0\}$ to represent scenarios with limited, moderate, and abundant privacy budgets, respectively.

\noindent \textbf{Baselines}. In this section, we mainly consider comparing \margnet's algorithm modules with those in state-of-the-art NN-based algorithm, \gem~\cite{liu2021iterative}, and propose two algorithms: 
\begin{enumerate}[leftmargin=*, label=\textbullet]
    \item \emph{\margnet-q}: We replace marginal feature by marginal query in \margnet, denoted as \margnet-q.
    \item \emph{\margnet-s}: We use a fixed-round iterative marginal selection strategy, which is the same as that in \gem, to select marginals and train them with neural networks.
\end{enumerate}

\noindent \textbf{Metrics}. To investigate the influence of algorithmic modules on utility, we prioritize utility-based evaluation. Consequently, our assessment focuses on machine learning efficacy, query error, and fidelity error.

We present the ablation study results in \Cref{tab: ablation}. First, marginals are more representative features than marginal queries. As analyzed in \Cref{sec: exist work}, marginals enable more efficient privacy budget utilization by measuring the frequency distribution over all unique values rather than a single scalar query. This helps \margnet achieve superior synthesis utility compared to \margnet-q. Second, adaptive selection proves more robust than the fixed-round iterative selection strategy, particularly under a limited privacy budget (e.g., $\varepsilon=0.2$). We attribute this to the fact that limited privacy budgets necessitate more judicious budget allocation, which adaptive selection provides through its dynamic termination criterion. This advantage persists on the Gauss10 dataset, but the performance gap narrows. Consistent with our findings in \Cref{sec: dense comparison}, we hypothesize that this is due to the coarse discretization imposed by PrivTree under a limited privacy budget, which introduces substantial discretization errors that overshadow the benefits of adaptive selection.

\subsection{\rev{High-order Marginal Evaluation}}

\rev{The preservation of high-order distributions is an evaluation metric frequently overlooked in prior studies. Historically, this metric has received limited attention because high-dimensional distributions are inherently sparse, and the computational cost of evaluating all such distributions is prohibitive. Consequently, we restrict our evaluation of high-dimensional marginals to the query error, providing a supplementary analysis to our primary experiments. The corresponding results are illustrated in \Cref{tab:high_way_query}.}

\rev{As shown in \Cref{tab:high_way_query}, our primary observation is that the results for 4- and 5-way marginal queries on both the sparsely correlated dataset (Adult) and the densely correlated dataset (PAMAP2) generally align with the trends observed in low-dimensional distribution metrics. Specifically, \aim maintains an advantage on the sparsely correlated dataset, whereas \margnet performs slightly better on the densely correlated dataset when provided with a sufficient privacy budget. However, the differences between AIM and MargNet remain small for these high-order workloads, so these results provide supporting rather than decisive evidence.}

{
    \begin{figure*}[t]
        \centering
        \includegraphics[width=\textwidth]{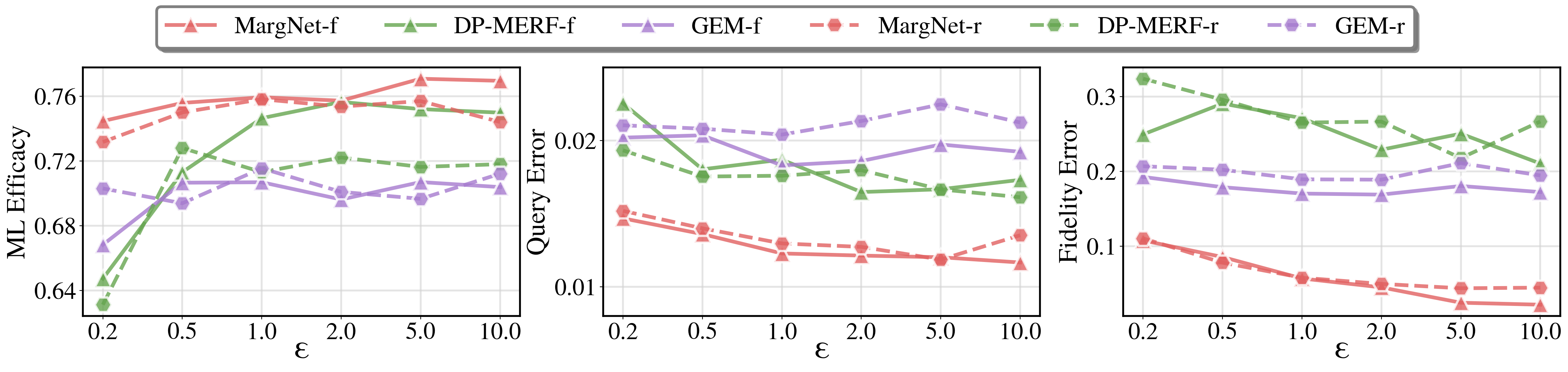}
        \caption{Machine learning efficacy, query errors, and fidelity errors of \margnet, \merf, \gem on Adult dataset. The real lines (``-f") mean methods with fixed input, and the dashed lines (``-r") mean methods with resampled input.}
        \label{fig: fix res}
    \end{figure*}
}

{
\begin{table}[t]
\footnotesize
\centering
\caption{Training hyperparameter ablation study results on the Adult dataset. We investigate three hyperparameters under $\varepsilon = 1.0$. Bold values are the default values in previous experiments.}
\label{tab: hyper ablation}
\resizebox{\linewidth}{!}{
\begin{tabular}{c|c|ccc}
\hline
  & {   Value   }  & { ML Eff. }   & { Query Err. }   & { Fidelity Err. }   \\
\hline
 \multirow{4}{*}{\makecell{Learning \\ Rate}}  
    & 1e-2 & 0.768 & 0.0122 & 0.0224\\
    & \textbf{1e-3} & 0.763 & 0.0117 & 0.0217\\
    & 1e-4 & 0.768 & 0.0112 & 0.0203\\
    & 1e-5 & 0.747 & 0.0119 & 0.0298\\
 \midrule
 \multirow{4}{*}{\makecell{Training \\ Iterations}}   
    & 50 & 0.762 & 0.0125 & 0.0209 \\
    & \textbf{100} & 0.763 & 0.0117 & 0.0217 \\
    & 200 & 0.764 & 0.0114 & 0.0219 \\
    & 400 & 0.767 & 0.0129 & 0.0232 \\
 \midrule
 \multirow{4}{*}{\makecell{Batch \\ Size}}  
    & 32 & 0.725 & 0.0127 & 0.0423 \\
    & 128 & 0.745 & 0.0129 & 0.0276 \\
    & \textbf{512} & 0.763 & 0.0117 & 0.0217 \\
    & 1024 & 0.768 & 0.0119 & 0.0218 \\
\hline
\end{tabular}
}
\end{table}
}

\subsection{Ablation Study of Hyperparameters}

In this subsection, we conduct some ablation experiments on training parameters (learning rate, training iterations, and batch size). The results are shown in \Cref{tab: hyper ablation}.

\noindent \textbf{Datasets}. For simplicity, we conduct experiments on $\varepsilon = 10.0$ and on the Adult dataset. 

\noindent \textbf{Baselines}. We mainly focus on three NN-based algorithms: \margnet, \merf, and \gem. \margnet-f, \merf-f, and \gem-f are these algorithms with fixed input, and \margnet-r, \merf-r, and \gem-r refers to these methods without fixed input.

\noindent \textbf{Metrics}. To investigate the impact of hyperparameters on algorithmic utility, we focus our evaluation on machine learning efficacy, query error, and fidelity error.

For the learning rate, when it decreases to a low level (e.g., 1e-5), we observe a slight drop in performance. We believe this occurs because the model cannot converge under such a small learning rate. For training iteration, we find that within a reasonable interval, it does not significantly influence the model performance. A notable hyperparameter is batch size, which in our analysis can affect the model's theoretical fitting ability. In the ablation study, when the batch size is small (e.g., $b = 32$), the performance degrades.

{
\begin{table}[t]
\footnotesize
\centering
\caption{Maximum memory cost of different algorithms. NN (all) and PGM (all) mean utilizing the neural network and PGM to fit all pairwise marginals, respectively. The results are obtained on the Adult dataset.}
\label{tab: memory}
\begin{tabularx}{\columnwidth}{l|XXXX}
    \toprule
    {  $\varepsilon$  } & {\margnet} & {NN (all)} & {\aim} & {PGM (all)} \\
    \midrule
    1.0 & 68 MB & 206 MB & 597 MB & $>$ 1 TB\\
    10.0 & 130 MB & 206 MB & 1.55 GB & $>$ 1 TB \\
    \bottomrule
\end{tabularx}
\end{table}
}

\subsection{Memory Comparison of \aim and \margnet} 

Another key factor we consider when selecting algorithms to generate data. This subsection will provide the memory cost of PGM and neural networks. We mainly consider three cases: NN with all marginals, NN in \margnet, PGM with all marginals, and PGM in \aim. Notice that \aim is a CPU algorithm, but neural networks require GPUs. We only count the peak memory cost on the respective devices. The results are provided in \Cref{tab: memory}.

Firstly, aligning with what we mentioned, when we require fitting all pairwise marginals, PGM consumes much memory ($>$ 1 TB). The phenomenon is caused by the extremely large clique, whose domain size is equal to that of the whole dataset. In contrast, even with all marginals, neural networks can still work with relatively lower memory costs (206 MB). Furthermore, although we separate the marginal case and directly compare \margnet and \aim, the PGM still requires more memory than \margnet. Finally, we also need to clarify that these results are only on the Adult dataset. When dealing with other large datasets with plenty of privacy budget, \margnet is likely to cost more memory. This is because \aim constraints the marginal candidates to control the model size and make optimization feasible, but \margnet does not set any limitation.

\subsection{Fixing Input}

When representing the dataset by features (e.g., marginals), our objective becomes producing a batch of outputs that preserve these features more effectively. Accordingly, if the features are fixed, increased output diversity does not contribute to achieving this objective. In this case, we hypothesize that using a fixed model input (e.g., sampling input once at initialization and fixing it as the input for any future model access) can be helpful in model training on feature-based models. In this section, we conduct experiments on whether fixing input noise can have a positive influence on those feature-DP methods:

\begin{enumerate}[leftmargin=*, label=\textbullet]
    \item We pre-compute random Fourier feature (used in \merf) and marginals (used in \margnet and a special case in \gem), and train the neural network on these features. The loss tracks are plotted in \Cref{fig: merf loss} and \Cref{fig: marginal loss}, respectively.

    \item We directly compare the utility performances of \margnet, \gem, and \merf with those of these algorithms with fixed input. The results are illustrated in \Cref{fig: fix res}.
\end{enumerate}

{
    \begin{figure}[!t]
        \centering
            
            \begin{subfigure}[b]{0.48\linewidth} 
                \centering
                \includegraphics[width=\linewidth]{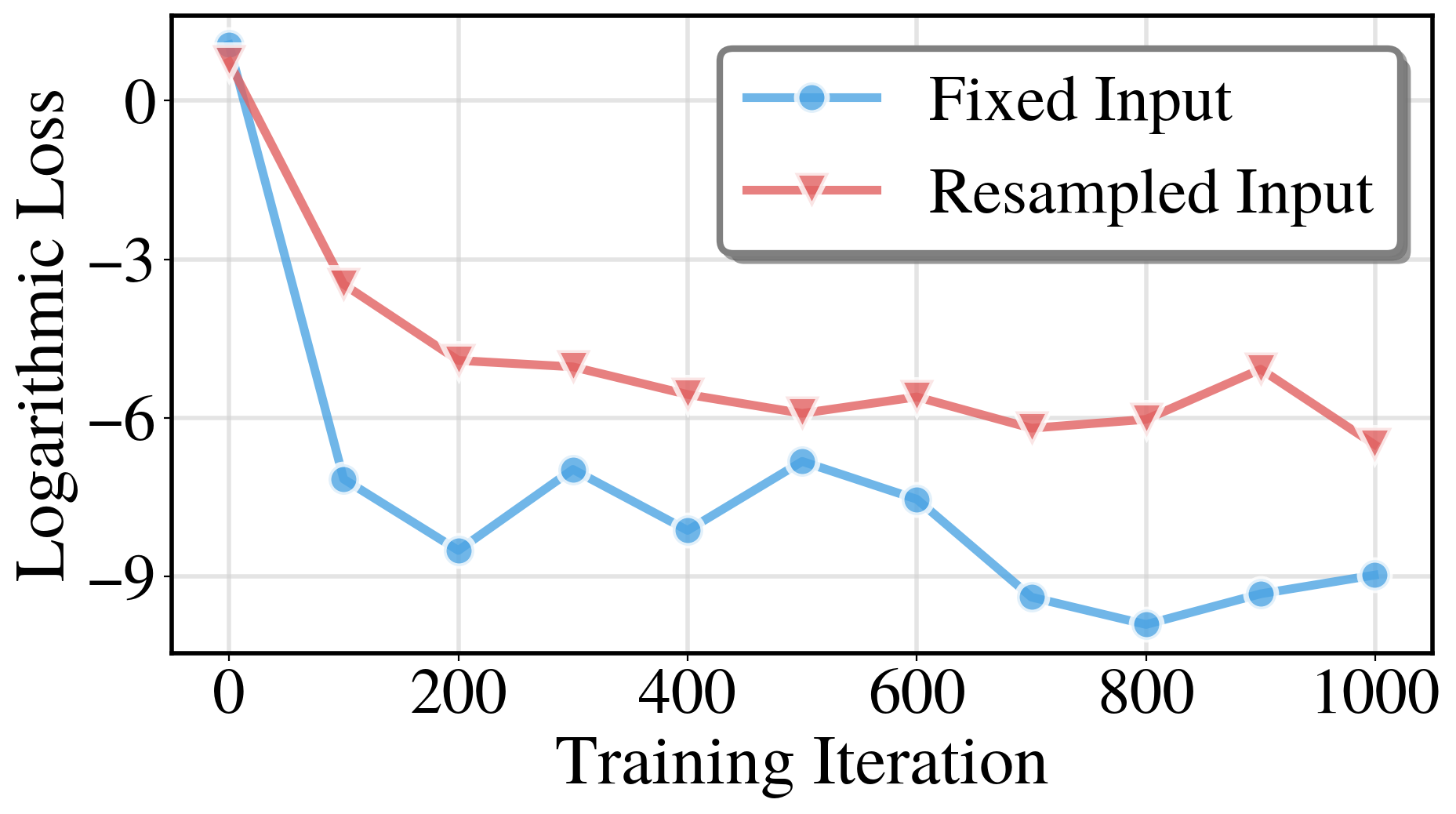}
                \caption{Random Fourier feature}
                \label{fig: merf loss}
            \end{subfigure}
            \hfill 
            \begin{subfigure}[b]{0.48\linewidth}
                \centering
                \includegraphics[width=\linewidth]{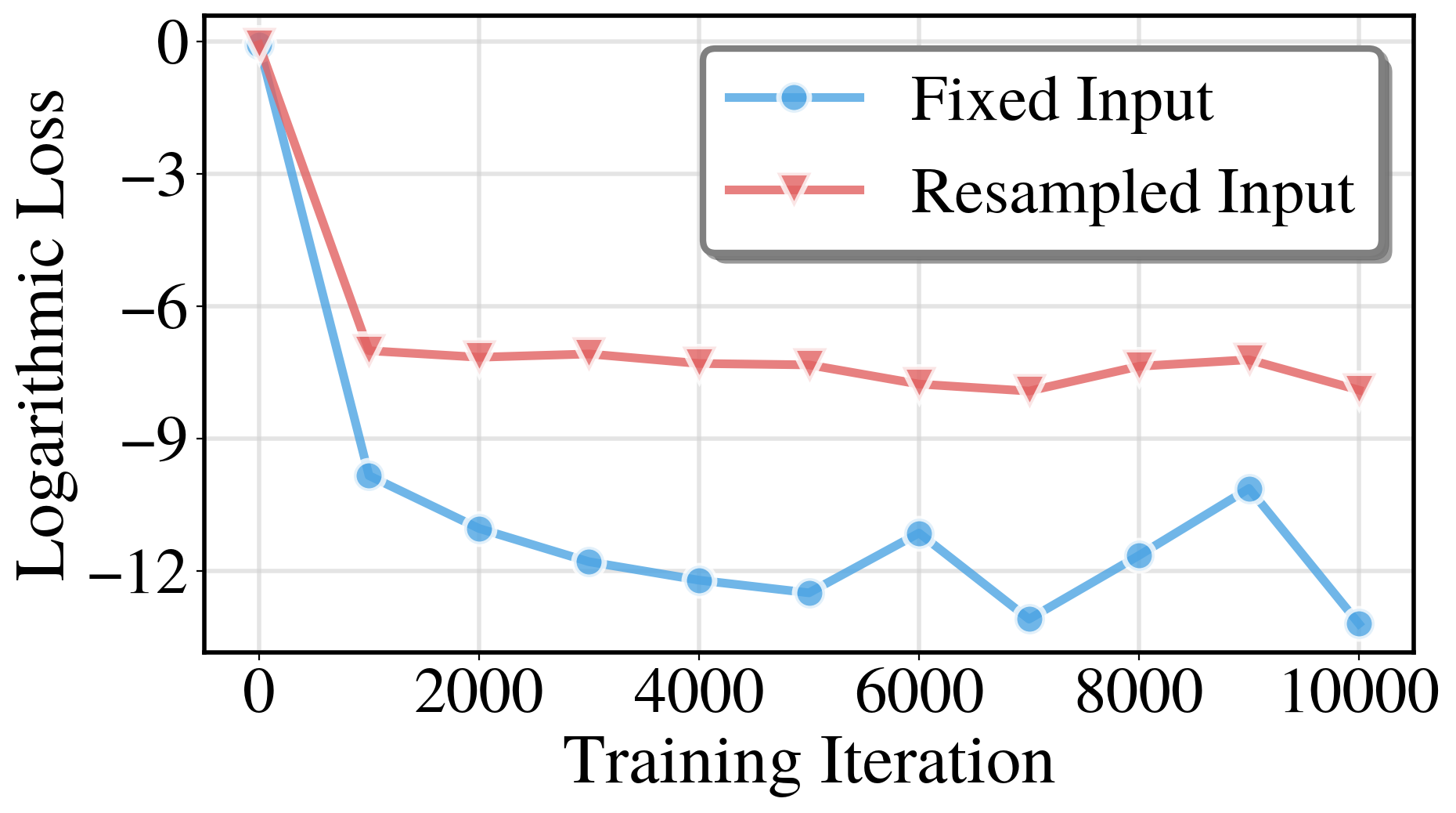}
                \caption{Marginal feature}
                \label{fig: marginal loss}
            \end{subfigure}
        
        \caption{Training loss under fixed and resampled input.}
        \label{fig: loss comparison}
    \end{figure}
}

In \Cref{fig: loss comparison}, a straightforward observation is that with fixed input, NN's training loss can converge to a lower level in limited training iterations. We believe this is because fixed input reduces the complexity of the map from input to output and helps models learn better. In \Cref{fig: fix res}, the performances of \margnet, \merf, and \gem under fixed input are not necessarily inferior to their counterparts under resampled input; in fact, they are sometimes superior. These results lend support to our conjecture that fixing the input has at least no detrimental effect on the model's training or final generative performance. Therefore, in our main experiments, we uniformly adopt this fixed-input setting for \margnet.